\documentclass[journal]{IEEEtran}
\usepackage{blindtext}
\usepackage{graphicx,color}
\usepackage{subfig}
\usepackage{algorithm}
\usepackage{mathrsfs}
\usepackage{multirow}

\usepackage[cmex10]{amsmath}
\usepackage{algorithmic}
\usepackage{soul}	
\ifCLASSINFOpdf
\else
\fi
\begin{document}
%
\title{Deep Class-Wise Hashing: \\ Semantics-Preserving Hashing via Class-wise Loss}
%
%

\author{Xuefei Zhe,~
        Shifeng Chen,~\IEEEmembership{Member,~IEEE,}
        and Hong Yan,~~\IEEEmembership{Fellow,~IEEE}
\thanks{Xuefei Zhe and Hong Yan are with the Department of Electronic Engineering, City University of Hong Kong, Hong Kong.}
\thanks{Shifeng Chen is with Shenzhen Institutes of Advanced Technology, CAS, China.}
\thanks{Corresponding author: Xuefei Zhe (e-mail: xfzhe2-c@my.cityu.edu.hk).} 
}

\maketitle

\begin{abstract}
Deep supervised hashing has emerged as an influential solution to large-scale semantic image retrieval problems in computer vision. 
In the light of recent progress, convolutional neural network based hashing methods typically seek pair-wise or triplet labels to conduct the similarity preserving learning. 
However, complex semantic concepts of visual contents are hard to capture by similar/dissimilar labels, which limits the retrieval performance.
Generally, pair-wise or triplet losses not only suffer from expensive training costs but also lack in extracting sufficient semantic information. 
In this regard, we propose a novel deep supervised hashing model to learn more compact class-level similarity preserving binary codes. 
Our deep learning based model is motivated by deep metric learning that directly takes semantic labels as supervised information in training and generates corresponding discriminant hashing code. 
Specifically, a novel cubic constraint loss function based on Gaussian distribution is proposed, which preserves semantic variations while penalizes the overlap part of different classes in the embedding space. 
To address the discrete optimization problem introduced by binary codes, a two-step optimization strategy is proposed to provide efficient training and avoid the problem of gradient vanishing. 
Extensive experiments on four large-scale benchmark databases show that our model can achieve the state-of-the-art retrieval performance. Moreover, when 
training samples are limited, our method surpasses other supervised deep hashing methods with non-negligible margins.
\end{abstract}


%
\IEEEpeerreviewmaketitle

\section{Introduction}
\label{sec:intro}
Fast development of internet technologies and recent advancements in hand-held smart devices have boosted the visual contents sharing. 
The convenient access to internet and interests in socializing platforms have flooded the image based information contents such as human faces, sceneries, online products, etc. 
Such an enormous growth of image data has paved the way for emerging applications from different domains where the users are interested in retrieving images from a large-scale collection. 
Remarkably little has yet been produced to meet the visual content based retrieval challenges effectively for large-scale data. 
Unlike traditional search engines where retrieval is inherently dependent on index terms and tags related to images, content based image retrieval (CBIR) manipulates large multimedia database by characterizing them based on their respective visual features.

Recent years have witnessed numerous achievements in CBIR~\cite{kulis2009fastCBIR,gong2013iterativeCBIR,li2014subCBIR}. 
In spite of large volumes of retrieval databases, the approximate nearest neighbor (ANN) searching methods have rendered satisfiable performances to capture visual contents  with moderate retrieval time. 
Among many ANN searching methods, hashing nearest neighbor methods remain at the top~\cite{wang2017survey}. 
The superior performance of hashing methods is credited to robust mapping functions that can project an instance, such as an image, into a compact binary code. 
The resultant binary codes have a shorter distance under the Hamming metric if their original instances are similar. 
In general, hashing methods are considered more effective than other ANN based methods regarding both searching speed and memory cost.

Existing hashing methods can be divided into data-independent and data-dependent ones by considering whether training data are used. 
Data-dependent methods are also known as learning to hashing (L2H). 
Taking benefits of the training data, L2H usually achieves much better performance compared with data-independent methods. 
Hence, the study of data-dependent L2H methods attracts more attention.
Data-dependent learning to hashing can be further categorized into unsupervised and supervised methods by considering whether the supervised information is used. 
More information about learning to hashing can be found in the survey paper \cite{wang2017survey}.

Among many supervised learning to hashing methods, deep supervised hashing, which is based on deep neural network structures, has shown an encouraging ongoing research progress, with the aim to learn an end-to-end structure that can directly map images into binary codes.
Different from deep hashing, the classical approach of supervised L2H is generally composed of two steps: firstly extracting off-the-shelf hand-crafted features such as HOG~\cite{HOG}, GIST~\cite{gist}, SIFT~\cite{SIFT} and variants of SIFT~\cite{ke2004pcasift,GSIFT}, as the input of the hashing learning procedure; then learning the hashing function based on these visual features. 
Obviously, for this kind of L2H methods, the final performance  is influenced not only by the mapping function, but also by the quality of used visual features. 
Previous research efforts on hashing learning  such as supervised discrete hashing (SDH) \cite{shen2015supervisedhashing}, fast supervised hashing \cite{lin2014fastH} and column sampling supervised hashing \cite{kang2016column} mostly focus on the second step. 
By taking feature learning into account, deep hashing involves the convolution neural network (CNN) structure as a feature learning part of the hashing learning framework. 
As a result, deep hashing usually gives better performance than classical  supervised L2H methods on the image retrieval task.  

To achieve better performance and training efficiency, the following two essential questions need to be fully exploited:
\begin{itemize}
	\item how to use supervised information to serve the similarity-preserving learning; 
	\item how to effectively solve the discrete optimization problem introduced by binary codes.
\end{itemize}

 Currently, the widely available supervised information of image data is the semantic label information.
 For example, CIFAR-10 \cite{CIFAR10} and ImageNet \cite{ILSVRC15} provide single class label for each image  while NUS-WIDE \cite{chua2009nus} and COCO \cite{coco} assign multiple labels to a single image. 
 Several methods generate similar or dissimilar pair-wise labels from these available class labels~\cite{DPSH,2017hashNet}. 
 That is, two images from the same class are assigned with the similar label, otherwise they are treated as dissimilar.
 Then the pair-wise or ranking (triplet) loss is used for training \cite{xia2014supervised,DPSH,2017hashNet}. 
 The pair-wise or ranking (triplet) loss implicitly shows how the supervised information is used for the similarity preserving learning. 
 However, it is very expensive to train with the pair-wise or triplet loss because the number of inputs is the number of training samples to the power of two for \cite{DPSH} or three for \cite{DTSH}, respectively.
 Besides, how to sample the training pairs or triplets significantly impacts the final performance \cite{facenet}.
 Another common solution is to directly use the softmax to guide the training process \cite{yang2017superviseddeepbinary}. 
 Training with softmax is much simpler compared to that with the pair-wise or ranking loss. 
 But how the softmax serves the similarity preserving learning is not fully understood. For example, softmax is used by combining with the pair-wise or the ranking labels in~\cite{yao2016}.

The other challenge comes from solving the discrete optimization with the binary constraint. 
Under the scenario of deep hashing, images are mapped to binary codes in the form $b\in \{-1,+1\}^n$. 
Works in \cite{DPSH,yansemi} directly move this discrete constraint to a regularization term that minimizes $\sum^n_{i=1}||sign(x_i) - x_i ||^2_2$, where $x_i$ is the continuous feature vector generated by the deep neural network. 
Though this relaxation is straightforward, it is unfeasible to train the neural network with standard back-propagation due to all zero-valued gradient of the $sign$ function for non-zero inputs \cite{2017hashNet}.
Another common approach introduces an active function, such as $sigmoid$ or $tanh$, \ to restrict output within $[0,1]$ or $[-1,1]$ as well as setting hyper parameters to push outputs close to the saturate parts \cite{2017hashNet,yang2017superviseddeepbinary}. 
Though this method eliminates the $sign$ function, it faces the vanishing gradient problem due to the usage of $sigmoid$ or $tanh$. 
Specifically, the outputs are forced to be close to the saturate part where gradients are extremely small.
This problem becomes non-ignorable when the CNN part has deeper structures. 
 
In this paper, we propose a novel deep supervised hashing method, which  effectively exploits the class label information. 
Different from previous deep supervised hashing that uses similar/dissimilar labels for training, our observation is that the class labels themselves are more natural and contain richer semantic concepts than the similar/dissimilar labels. 
Meanwhile, directly using the class labels is also beneficial for training since training with pair-wise or triplet loss function, which is generated by class labels, is computationally being expensive \cite{MagnetRippel2016,zhuang2016fasttriplet}.
%
Based on these observations, we develop a novel loss function which directly uses the class labels to serve the similarity-preserving learning by punishing the overlap part among different classes in the embedding space. 
Because our model is designed to handle the deep supervised hashing through class wise level, we call it Deep Class-Wise Hashing (DCWH). 
Besides, we extend our model to multi-label data. 

On the other hand, instead of directly solving the discrete optimization problem, we suggest a two-stage optimization strategy that guarantees our model can be effectively trained. 
From a high level point of view, we first solve an approximate version of the optimization problem with the hyper cube constraint. Instead of directly learning a discrete feature space, we propose first to project images into a hyper cube space, which is more suitable for the continuous nature of the CNN.
Then, at the second stage, we continue refining our model to reduce the quantization error. 
Extensive experiments on four import benchmark datasets show that our model surpasses other state-of-the-art deep hashing methods.

To summarize, the main contributions of this paper are as follows:
\begin{itemize}

	\item A novel loss function based on the Gaussian model is proposed to handle the deep hashing learning problem at the class level. By directly employing class labels, the new objective function is used to reduce the intra-class spread mean while enlarge the inter-class gap.
	\item A two-stage strategy is presented to optimize the discrete objective function for DCWH. 
	Specifically, the problem is firstly optimized with the cubic constraint as a sub-optimal problem and then continues to be refined with the discrete constraint.
	\item Experiments on four important benchmark datasets show DCWH can outperform other state-of-the-art deep hashing methods. 
\end{itemize}

The rest of this paper is organized as follows. 
Sec~\ref{Sec:relatedworks} briefly introduces recent research works in deep hashing and deep metric learning.
The motivation of how we develop the class-wise model from the pair-wise loss is explained in Sec~\ref{Sec:motivation}. 
Based on the motivation, a class-wise deep hashing model along with the optimization strategy is presented in Sec~\ref{Sec:model}. 
In Sec~\ref{Sec:experiments}, extensive experiments is conducted to compare with other state-of-the-art deep hashing methods. 
In Sec~\ref{Sec:discussion}, several important properties of the proposed model are discussed. 
The conclusion and feature works of this paper are drawn in the final section.

\section{Related Works}
\label{Sec:relatedworks}
In this section, several deep supervised hashing methods are reviewed. 
Then, considering that deep hashing can be treated as a special case of deep metric learning, we briefly discuss some latest deep metric learning approaches. 

\subsection{Deep Supervised Hashing}
Recent deep hashing methods combined feature learning and hashing learning into an end-to-end learning system. 
The early work, Convolutional Neural Network Hashing (CNNH)~\cite{xia2014supervised} built a two-stage hashing learning method. 
The hashing codes are learned with pair-wise labels in the first stage. 
At the second stage, the hashing codes are used to learn feature representation. 
One main drawback of CNNH is that the representation learning part (the second stage) does not give feedback to hashing learn part. 
To overcome the above issue, deep pairwise-supervised hashing ~\cite{DPSH} and HashNet~\cite{2017hashNet} proposed different end-to-end hashing learning approaches based on pair-wise labels. 
Besides, DNNH~\cite{lai2015DNNH} utilized the triplet ranking loss to guide the hashing learning . 
However, training with pair-wise or triplet loss is time-consuming because of following reasons. 
First, the extra processing is needed to generate mini-batch with the pair or triplet structure ~\cite{zhuang2016fasttriplet,MagnetRippel2016}. 
Second, the number of total samples is increased exponentially.
On one hand, not all pairs or triplets can provide useful information. 
On the other hand, the data imbalance problem occurs because there are much more dissimilar pairs than similar ones~\cite{2017hashNet}.
Another shortcoming attracting much attention recently is that pair-wise or triplet loss does not fully use the semantic (class) information \cite{DSDH}. 
Currently, the most available training data are class-label based data. 
The pair-wise or triplet labels are generated from class labels.
However the semantic information cannot be fully presented in such format. 

To better use the semantic label information, SSDH \cite{SSDH} directly uses a softmax classifier to guide the deep hashing learning. 
DSRH \cite{yao2016DSRH} proposes a method that utilizes triplet loss hashing stream and classifier stream to perform as multi-task learning.
However, the classifier stream does not impact the hashing code learning. 
Deep supervised Discrete Hashing~\cite{DSDH} considers that learned hashing codes should be ideal for classification. 
So, in this approach, a classifier loss layer is added to the hashing layer and co-trained with the pair-wise loss. 

\subsection{Deep Metric Learning}
Many existing deep hashing methods are based on the pair-wise or the triplet loss, which is also used in deep metric learning.
This indicates that deep hashing can be treated as a special case of deep metric learning, of which the embedding space is a Hamming space. 
Recent deep metric learning has achieved many improvements in both performance and training efficiency. 

Deep metric learning aims to learn a non-linear projection function which can transform an image from pixel level to a discriminate space where samples from the same class will be gathered together, and samples from different classes will be pushed apart. 
Different from the classical distance metric learning which only optimizes the distance metric function based on existing features, e.g., hand-craft features, the deep metric learning directly learns from images and generally can achieve better performance. 
The pairwise and triplet loss are widely used in the deep metric learning frameworks and have been successfully applied to many visual tasks, for example, fine-grained categorization \cite{cui2016fine}, image retrieval and deep hashing learning \cite{DPSH}, face verification \cite{facenet}, and person re-identification \cite{personReID}. 
The performance of triplet-based deep metric learning highly relies on the quality of triplet pairs. Many methods have been proposed to deal with this, for example, mining pairs with rich information \cite{facenet} or containing more pairs within a mini-batch ~\cite{QuintupletHuang2016,liftedStructure}. 
The improvement comes along with a more complicated training process. 
Different from considering adding more pairs, the Magnet~\cite{MagnetRippel2016} proposed a method that punishes the overlap between different clusters. 
Though the convergence is shown to be faster than triplet based methods, it is still very time consuming for generating mini-batches by retrieving images from adjacent clusters.

\begin{figure*}[t!]
	\centering
	\includegraphics[width=0.8\textwidth]{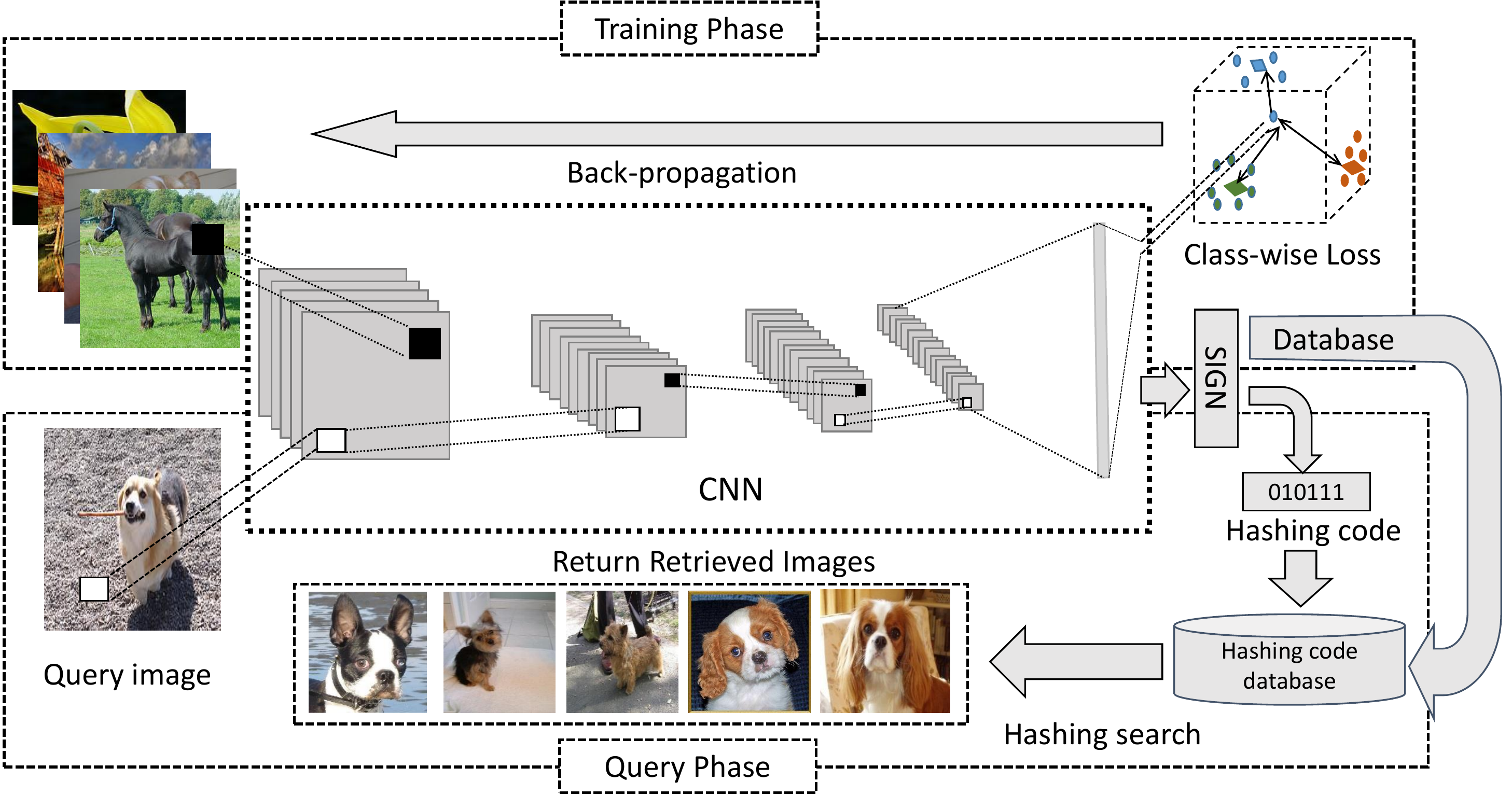}
	\caption{The end-to-end deep hashing framework of DCWH. The center part is a CNN structure shared by both the training and the query phases.}
	\label{fig:system}
\end{figure*}
\section{Motivation: From Pair-Wise to Class-wise}
\label{Sec:motivation}
In this section, we show the motivation how our class-wise loss is developed from pair-wise loss.
The pair-wise \cite{DPSH,yang2017pairwise} and triplet loss \cite{lai2015DNNH,zhu2016DHN,zhuang2016fasttriplet} are two widely used approaches for deep supervised hashing. 
The main idea of pair-wise method is to decrease the distance between binary codes of similar instances and increase the distance between dissimilar ones. 
Given a anchor image $X_a$, the objective function of pair-wise can be simply formulated as follows:
\begin{equation}
    \underset{\theta}{\text{min}}~ \sum_{i=1}^{N}~ \{S_{a,i}{d}(f_{\theta}(X_a),f_{\theta}(X_i))  - (1 - S_{a,i}){d}(f_{\theta}(X_a),f_{\theta}(X_i))\},    
\end{equation}
where $X_i$ is an input instance, $d$ is a distance metric function and $\theta$ are parameters in the hashing function $f$. 
$S_{a,i}$ is a pair-wise label, which equals to $1$ when the image $a$ and the  image $i$ are similar, otherwise $S_{a,i}$ equals to $0$. 
Instead of only considering the relation between two instances, the triplet loss considers the distance within a triplet structure. 
That is, for an anchor instance, the distance to a dissimilar instance should be larger than the distance to a similar one with a margin $m$, which can be treated as a double-pair version of pair-wise loss. 
Given an anchor image $X_i^a$, a negative (dissimilar) one, denoted as $X_i^n$ and a positive (similar) one, $X_i^p$, are selected to form a training sample. 
The hinge loss function of the triplet is presented as follows:
\begin{equation}
    \underset{\theta}{\text{min}}~ \sum_{i=1}^N~   \max({d}(f_{\theta}(X_i^a),f_{\theta}(X_i^p))-{d}(f_{\theta}(X_i^a),f_{\theta}(X_i^n)) + m~,0).    
\end{equation}

Though the total number of training samples is increased, not all triplets or pairs can provide useful information. 
Triplet selection \cite{facenet} or hard negative example mining methods are proposed to deal this problem. 
As a side product, the training procedure becomes more complicated. 
Latest deep metric learning research works improve performance and training efficiency by constructing more data relations within one mini-batch. 
Work in~\cite{liftedStructure} considered constructing more pairs to make full use of images in one batch.
Quintuplet loss~\cite{QuintupletHuang2016} introduced a quintuplet loss which provides three pairs for a single anchor image from both cluster and class levels. 
By taking the cluster structure within one class into account, the Magnet~\cite{MagnetRippel2016} proposed a method to punish overlap between different clusters.
A similar point shared by above improvement methods is to construct more pairs for the anchor images. 
This observation leads us to think whether the performance can be further improved by adding more pairs. 
An extreme case would be that, given an anchor image, all similar and dissimilar images from the whole training set are provided to calculate the loss. 
This idea can be formulated as follows:
\begin{equation}
   \underset{\theta}{\text{min}} \sum_{i=1}^N   \{ \sum_{j=1}^{N_{is}} {d}(f_{\theta}(X_i^a),f_{\theta}(X_j))  - \sum\limits_{k=1}^{N_{id}}{d}(f_{\theta}(X_i^a),f_{\theta}(X_k))\},
\end{equation}
where $N_{is}$ is the total number of similar images to the anchor image, $X_i^a$. $N_{id}$ equals to the total number of dissimilar ones to $X_i^a$. 
Though this extreme case can provide better global information, it is not practical for training a CNN. 
Because only a few images can be handled within a mini-batch. 
Rethinking that similar images defined in semantic retrieval are the images with the same class label. We rewrite above loss function as:
\begin{equation}
\begin{split}
   \underset{\theta}{\text{min}}~ \sum_{i=1}^N~  & \left\{ \sum_{j=1}^{N_{C_a}} {d}(f_{\theta}(X_i^a),f_{\theta}(X_j)) \right. \\
                                             & \left. - \sum_{C\neq C_a}\sum\limits_{k=1}^{N_{C}}{d}(f_{\theta}(X_i^a),f_{\theta}(X_k)) \right\},
\end{split}
\end{equation}
where $N_C$ indicates the number of images in class $C$. 
$C_a$ is the class label of the anchor image $X_i^a$. 
If $d$ is an Euclidean metric, the above optimization problem has the same solution as:
\begin{equation}
\begin{split}
   \underset{\theta}{\text{min}}~ \sum_{i=1}^N~  & \left\{||f_{\theta}(X_i^a) - \frac{1}{N_{C_a}}\sum_{j=1}^{N_{C_a}}f_{\theta}(X_j)|| \right. \\
                                             & \left. - \sum_{C\neq C_a}||f_{\theta}(X_i^a) - \frac{1}{N_{C}}\sum\limits_{k=1}^{N_{C}}f_{\theta}(X_k)|| \right\}.
\end{split}
\end{equation}
Denote $\mu_C  =\frac{1}{N_{C}}\sum_{j=0}^{N_{C}}f_{\theta}(X_j)$ as the center of class $C$. A simple formulation is:
\begin{equation}
\label{equ:1}
   \underset{\theta}{\text{min}}~ \sum_{i=1}^N~   \left\{||f_{\theta}(X_i^a) - \mu_{C_a}|| 
                                              - \sum_{C\neq C_a}||f_{\theta}(X_i^a) - \mu_C|| \right\},
\end{equation}
For now, Equation~(\ref{equ:1}) is more practical for training a neural network. 
The remain question is how to compute the class centers. 
Considering that $\theta$ is optimized with stochastic gradient descent based on the mini-batch, the class centers do not change dramatically within limited iterations. 
A simple idea is that the class centers can be periodically updated.
In the following part, we will formulate our class-wise deep hashing learning.
\section{\textsc{Deep Class-wise Hashing}}
\label{Sec:model}
In this section, a system overview of our model will be given first. 
Then Equation~(\ref{equ:1}) will be formulated as a class-wise deep hashing loss followed by a novel optimization strategy. 
At the end of this section, the extension to multi-label data will be given.
\subsection{System Overview}
Figure \ref{fig:system} provides the overview of our system. The whole system has two phases: training and query. 
Both phases share the same convolutional neural network. 
For the training phase, the CNN as a feature learning part is trained with our class-wise loss. 
After training, an element $sign$ layer is added to generate hashing codes. 
By forwarding the database images, a hashing code database is built.
For the query phase, when receiving a query request, the system produces the query hashing code by forwarding the query image, then finds the top nearest images via searching in the hashing code database.
\subsection{Problem Definition}
Given a training set of $N$ labeled samples, denoted as $\{x_n,y_n\}_{n=1}^N$ belonging to $C$ classes, our goal is to learn a transformation function denoted as $f(x_n; \Theta)$ where $\Theta$ represents all parameters in the transformation function and $x_n$ is input data. 
Specifically, a convolutional neural network is chosen as the hashing function in our system. 
By using this transform function, the original data $\{x_n\}$ are projected to a point $r_n \in \{0,1\}^L$ in a Hamming space with dimension $L$, where the intra-class distance is smaller than the inter-class distance. 
In other words, samples from the same class is closer to each other than samples from different classes under the Hamming metric.
\subsection{Objective Function}
Instead of directly using the Euclidean metric in Equation (\ref{equ:1}), we make use of a normalized probability model based on the Gaussian distribution to learn a compact feature space:
\begin{equation}
P(y_n|r_n^{(m)}£©;\Theta)=\frac{\frac{1}{\sqrt{2\pi}\sigma_m}\exp\{-\frac{1}{2 \sigma^2_m } D^2(r_n^{(m)} ,\mu_{y_n})\}}{\sum_{i=1}^{C} \frac{1}{\sqrt{2\pi}\sigma_i}\exp\{-\frac{1}{2 \sigma^2_i} D^2(r_n^{(m)} ，{\mu}_{i})\}} ,
\end{equation}
where $r_n^{(m)} = f(x_n^{(m)}; \Theta)$ is a feature vector obtained from the feature learning part, and $x_n^{m}$ represents a sample with the class label $y_n=m$. 
Class centers $\{\mu^{i}\}_1^C$ are updated using training data periodically. 
Different $\{\sigma_i^2\}$ can be used to capture variances of different classes. $\sigma$ can be adaptively updated \cite{MagnetRippel2016} or treated as learnable parameters \cite{ioffe2015batch}. 
In this paper, all $\{\sigma_i\}$ values are simply set to a fixed value as a hyper parameter, which performs as a global scaling factor to control class gaps. 
The function $D$ is a distance function. 
This likelihood function can be interpreted as the normalized probability assigned to the correct label $y_d = m$ given feature vector $r_n^{(m)}$, $M=\{\mu_i\}$ and parameters $\Theta$ of function $f$. 
By taking negative log-likelihood of the whole given data set $\{x_n, y_n\}$, the optimization problem can be formulated as follows:
\begin{equation}
\begin{split}
\label{eq:loss}
\underset{\Theta, \mathcal{M}}{\text{min}} ~\mathrm{J}=&-\log ~P(Y|X;\Theta,M)\\
=& -  \sum_{n=0}^N \log ~P(y_n|x_n^{(m)}£©;\Theta,M)\\
= & - \sum_{n=0}^{N} \log  \frac{\exp\{-\frac{1}{2 \sigma^2 } D^2(r_n^{(m)} ,\mu_{y_n})\}}{\sum_{i=0}^{C} \exp\{-\frac{1}{2 \sigma^2} D^2(r_n^{(m)} -{\mu}_{i})\}}.
\end{split}
\end{equation}
Notice that $r_n \in \{-1,1\}^L$ and $\mu_i \in \{-1,1\}^L$ are binary codes with $L$ bits. 
Since the class label information is directly used in our approach, we name our method as deep class-wise hashing (DCWH).

\subsection{Optimization}
\begin{figure}[ht]
\centering
\subfloat[Stage I: cubic constraint]{\includegraphics[width=0.235\textwidth]{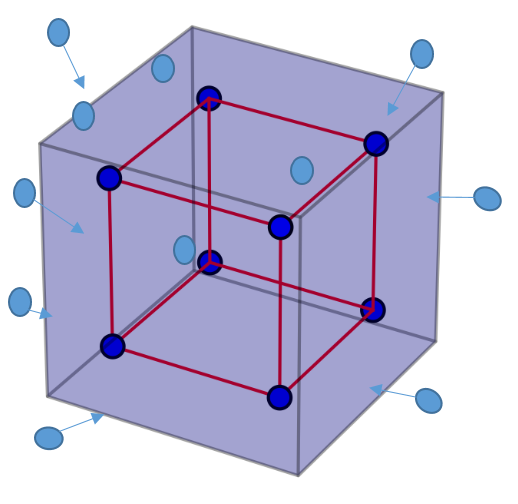}\label{fig:stage_1}}
\hfill
\subfloat[Stage II: vertex constraint]{\includegraphics[width=0.235\textwidth]{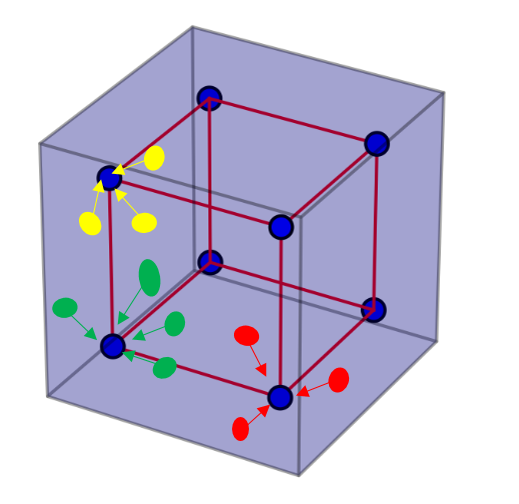}\label{fig:stage_2}}
\caption{Geometric interpretation of our two-stage strategy: at the first stage (a), dragging projected points into a hyper cube which is a little larger than hamming cube; and at the second stage (b), pushing projected points to vertexes (dark blue points) of the Hamming cube. {\textbf{Better viewed in color}}.}
\label{fig:two-stage}
\end{figure}

Due to the binary constraint, the proposed objective function is a discrete optimization problem. 
DPSH \cite{DPSH} and SSDH \cite{yang2017superviseddeepbinary} propose different relaxation strategies to handle this optimization problem. 
To avoid directly using $sign$ function, we propose a two-step strategy. 
At the first stage, we relax $r_n^{(m)}$ to a hyper cube. 
Then, at the second stage, we continue refining our model with the vertex constraint, which pushes points to the vertexes of the Hamming cube. 
This constraint is also known as the quantization error. 
At the first stage, projected points $\{r_n\}$ are constrained to move into a hyper cube which is a little larger than the Hamming cube ( a hyper cube with unit length). 
At the second stage, projected points $\{r_n\}$ are pushed to the vertexes of the Hamming cube. 
A geometric illustration can be found in Figure \ref{fig:two-stage}.
\subsubsection{Stage I: Establishing with Cubic Constraint}
To solve the optimization problem in Equation~(\ref{eq:loss}),  the following sub-optimal problem is considerd first:
\begin{equation}
\label{eq:st1}
\begin{aligned}
&\underset{\Theta,M}{\text{min}} & &\mathcal{J}_1=  -\sum_{n=1}^N \log\frac{\exp\{-\frac{1}{2 \sigma^2 } \|r_n^{(m)} -{\mu}_{y_n}\|^2_2\}}{\sum_{i=1}^{C} \exp\{-\frac{1}{2 \sigma^2} \|r_n^{(m)} -{\mu}_{i}\|^2_2\}}\\
& \textrm{subject to} & & r_n \in [-\alpha,\alpha]^L,\quad n =1,2,3,\dots,N, \\
&  & & \mu_i \in [-\alpha,\alpha]^L, \quad i = 1,2,3,\dots,C,
\end{aligned}
\end{equation}
where $M=\{{\mu_i}\}_{i=1}^C$ and $C$ is number of the total classes, $r_n^{(m)} = f(x_n^{m} ;\Theta)$, and $y_n = m$. $\alpha$ and $sigma$ are hyper parameters. 
$\alpha$ is set to be the length of the hyper cube, which should be slightly larger than the length of the Hamming cube. 
According to this requirement, we set $\alpha$ set to $1.1$. $L$ is the length of the hashing code.
By adding the cubic constraint as a regularization term to Equation~(\ref{eq:st1}), above problem can be formulated as follows:
\begin{equation}
\label{eq:updateclass}
\begin{aligned}
\underset{\Theta,M}{\text{min}} \quad \mathcal{J}_2 =& \mathcal{J}_1 +\eta_1 \left\{ReLU(-\alpha-r_n^{(m)})\right. \\
&	\left. +ReLU(r_n^{(m)}-\alpha) \right\},
\end{aligned}
\end{equation}
where $\eta_1$ is a regulation term. ${ReLU}$ is the rectified linear unit defined as $ReLU(x) = \max(0,x)$. $\{\mu_i\}$ is periodically updated as follows:
\begin{equation}
\mu_i = \frac{1}{n_i}\sum_{n=1}^{n_i}r_n^{(i)},
\end{equation}
where $n_i$ is total number of samples from class $i$. 
Since $\mathcal{J}_2$ is differentiable, we can back propagate the derivative of this loss during CNN training, and parameters $\Theta$ will be optimized. 
After converging, we can obtain optimized parameters $\Theta_1$ and class centers $M_1 = \{\mu_i\}$.

\subsubsection{Stage II: Refine with Vertex Constraint}
At the second stage, we will push projected points to vertexes of the Hamming cube. 
After obtaining $\Theta_1 $ from the first stage as the initialization, the original optimization can be considered as follows:
$$
\begin{aligned}
&\underset{\Theta,M}{\text{min}} & &\mathcal{J}_3=  -\sum_{n=1}^N \log~ \frac{\exp\{-\frac{1}{2 \sigma^2 } \|r_n^{(m)} -{\mu}_{y_n}\|^2_2\}}{\sum_{i=1}^{C}\exp\{-\frac{1}{2 \sigma^2} \|r_n^{(m)} -{\mu}_{i}\|^2_2\}}\\
& \textrm{subject to} & & r_n \in \{-1,1\}^L,\quad n = 1,2,3,\dots,N;\\
&  & &  {\mu_i} \in [-1,1]^L,\quad i=1,2,3,\dots,C.
\end{aligned}
$$
To solve this discrete problem, we can follow the the solution suggested in \cite{DPSH}. We set $\textbf{b}_n = sign(\textbf{r}_n^{(m)})$ and relax $r_n^{(m)}$ to continuous variables. 
By adding the vertex constraint term, the discrete problem becomes the following continuous one:
\begin{equation}
\begin{aligned}
&\underset{\Theta,M}{\text{min}} & &\mathcal{J}_4 =  \mathcal{J}_3 + \eta_2\sum_{d=1}^{N}\|b_n -r_n\|_2^2,\\
& \textrm{subject to} & & b_n \in \{-1,1\}^L, \quad n = 1,2,3,\dots,N,\\
& &&\mu_i \in [-\alpha,\alpha]^L,\quad i = 1,2,3,\dots,C,\\
\end{aligned}
\end{equation}
where $\eta_2$ is a regularization parameter. The $\{\mu\}_1^C$ can be also updated according to Equation~(\ref{eq:updateclass}) . 
If binary class centers ${\mu}_i \in \{-1,+1\}^L$ are needed, the updating can be done as \cite{HashClusteringong2015}. 
For each class, we compute the binary representation from CNN as $b_n^{m} =sign(r_n^{m})$. 
Then we compute the sum for each class as $s_m = \sum_{n=1}^{n_m} b_n^{m} $. 
Finally, the binary class center $\mu_i$ can be obtained as follows:
\begin{equation}
\mu_{mk} = sign(s_{mk}) =
						\begin{cases}
							+1, & \quad \text{if } s_{mk} \geq 0\\
                            -1, & \quad \text{if } s_{mk} < 0
						\end{cases},
\end{equation}
where $k$ is the vector entry and $k=1,2,3,\dots,L$. This process is similar to voting for every bit of the binary center within a class. 
Above objective function is also differentiable, which means that we can easily learn parameters of the CNN part using the back propagation method. 
The refining stage shares the same learning framework as Stage I, which is illustrated later.

The above two stages share the same learning framework. For each stage, two types of parameters need to be optimized: parameters of neural network $\Theta$ and class centers $M = \{\mu_i|i=1,2,3,\dots,C\}$.  We propose a simple alternative learning algorithm as follows:

\begin{algorithm}
\caption{ Learning algorithm for DCWH}\label{alg:1}
\begin{algorithmic}[1]

\STATE\textbf{Initialize} CNN parameters $\Theta$
\REPEAT
	\STATE for input samples, compute features $\{r_n^{(m)}\}$
    \STATE using feature $\{r_n^{m}\}$ to update class centers $\{\mu_i\}$
    \STATE using $\{u_i\}$ to calculate the loss function and perform back propagation to optimize $\Theta$ for N iterations
\UNTIL{convergence}
\end{algorithmic}
\end{algorithm}

For Stage I, we use parameters of the pre-trained model to initialize the CNN. 
For Stage II, parameters learned in stage 1 are used to initialize the CNN and a small regular $\eta_2$ is set for training.

\subsection{Hashing Codes Generation}
After completing the learning procedure, we can use the trained CNN to generate the hashing code by replacing our loss layer with an element-wise $sign$ function layer. 
This process is illustrated in Figure \ref{fig:system} as the query phase.

\subsection{Extend to the Multi-label Case}
Different from single label data, a multi-label instance is usually assigned to one or several labels or tags. 
A label vector $\mathbf{l}$ is used to represent the assigned labels. 
If the $c$-th label is assigned to data, $l_c =1$, otherwise, $l_c = 0$. 
Given an instance $x_n$ with the label vector $l_n$, the combination of assigned classes can be treated as a new semantic center. 
For example, if an image is denoted with ``people", ``dog" and ``cat", the combination of $\{\text{people, dog, cat}\}$ is considered as a new semantic center. 
The new semantic center $\hat{\mu}_n$ for the instance $x_n$ can be obtained as follows:
 \[
 \hat{\mu}_n =\frac{1}{\sum l_{ni}} \sum_{i=1}^C l_{ni} \mu_i, 
 \]
 where $i$ is the vector entry of ${l}_n$ and $C$ is the total number of classes.
  $\mu_i$ is the original class center of class $i$. 
The loss function of multi-label case then is formulated as follows:
\begin{IEEEeqnarray}{llll}
&\underset{\Theta, \mathcal{M}}{\text{min}} ~\mathrm{J} \nonumber\\
=&-\log ~P(Y|X;\Theta,M) \nonumber\\
=& -  \sum_{n=1}^N \log ~P(y_n|r_n;\Theta,M)\\
= & - \sum_{n=1}^{N} \log  \frac{h(r_n)}
{\sum_{i=1}^{C} (1-l_{ni}) \exp\{-\frac{1}{2 \sigma^2} D^2(r_n ,{\mu}_{i})\}+h(r_n)} , \nonumber
\end{IEEEeqnarray}
where $h(r_n) =\exp\{-\frac{1}{2 \sigma^2} D^2(r_n,~ \hat{\mu}_{n})\}$.
The new semantic centers $\hat{\mu}_n$ are generated during training. 
The original class centers are updated as the single label case but with a weighted mean manner:
\begin{equation}
    \mu_i = \frac{1}{\sum\limits_{n=1}^N l_{ni}} \sum\limits_{n=1}^N 
    \frac{l_{ni}}{\sum\limits_{i=1}^c l_{ni}}r_n .
\end{equation}
\subsection{Comparison with Magnet}
Among existing deep metric learning methods, our approach has similar objective formulation as Magnet \cite{MagnetRippel2016}. 
But there is still some important difference between the two methods. 
The loss function of Magnet can be simply written as:
\begin{IEEEeqnarray}{l}
\mathscr{L}=\frac{1}{MD}\sum^{M}_{m=1}\sum^{D}_{d=1}\nonumber\\
\left\{\log\frac{\exp\{-\frac{1}{2\sigma^2}\|r_d^m -\hat{\mu}_m\|^2_2\}}{\sum\limits_{\hat{\mu}:C(\hat{\mu})\neq C(r_d^m)}\exp\{-\frac{1}{2\sigma^2}\|r_d^m -\hat{\mu}\|^2_2\}}\right\},
\end{IEEEeqnarray}
where $r_d^m$ is a feature vector belong to cluster $m$. The cluster center, $\hat{\mu}_m$, is estimated based on each batch data as follows:
\[
\hat{\mu}_m=\frac{1}{D}\sum_{d=1}^D r_d^m
\]
The first difference is how to generate mini-batches. 
Our model does not rely on specific mini-batch structure, which is as simple as training with the softmax loss. 
To generate a mini-batch for Magnet, one cluster is first randomly selected, then $M-1$ closest clusters will be chosen. 
$D$ images per selected cluster will be randomly selected ($M\times D$ in total) to form a mini-batch. 
This process is very time-consuming. 
Secondly, our model set $\sigma$ as a hyper parameter and updating class centers by forwarding training set, both of which are estimated within a mini-batch for Magnet. 
Another main difference is that the loss function for Magnet does not include the numerator term in the the denominator. 
By adding numerator to denominator, our model can maintain inter-class divergence and perform as a normalized probability. 
\section{Experiment}
\label{Sec:experiments}
In this section, extensive experiments are conducted to compare our method with several state-of-the-art deep hashing methods. 
Two public single-label datasets are used to evaluate our model. 
Two multi-label datasets are used to verify the effectiveness of our multi-label extension. 
We divide our experiments into two groups to follow different experiments setting in DSDH~\cite{DSDH} and HashNet~\cite{2017hashNet}. 
For experiments in Group I, all images in the database are used as training samples. 
For Gourp II, the deep hashing methods are only trained with $1,0000$ images selected from the retrieval database. 
To conduct a fair comparison, the results with the same data setting are directly cited from the original paper. 
For the single label datasets, images from the same class are treated as similar. 
For the multi-label datasets, images shared at least one same label are considered as similar. 
The mean average precision (MAP) is used to evaluate the retrieval performance.

\subsection{Implementation Details}
We develop our DCWH model based on MXNet~\cite{chen2015mxnet}, an open source deep learning framework. 
Since the convolutional neural network structure is not our focus, we simply apply the GoogLeNet with the batch normalization~\cite{ioffe2015batch} as our feature learning part. 
The pre-trained model on the ImageNet is used to initialize the CNN part in Stage I. 
The final fully connected layer is initialized by the Xavier with the magnitude of $1$. 
We use the mini-batch stochastic gradient descent with learn-rate $0.001$, and $0.0005$ weight decay. 
All images are first resized to $224\times224$ pixels then fed for training. 
Our experiments run on a single Nvidia GTX-1080 GPU server with 64 images per mini-batch. 
$\eta_1$ and $\eta_2$ are set to $10$ and $0.01$ respectively. 
$\sigma^2$  for multi-label extension is fixed to $1$. 
The value of $\sigma^2$ for the single label model depends on the code length. 
The values of $\sigma^2$ for the single label model are presented in Table\ref{tab:sigma_bits} by the standard cross-validation procedure.
\begin{table}[h]
\caption{$\sigma^2$ \textsc{for different bits}}
\label{tab:sigma_bits}
\centering
\resizebox{0.35\textwidth}{!}{%
\begin{tabular}{|c|c|c|c|c|c|c|}	
\hline
bits & 12 & 16 & 24 & 32 & 48 & 64 \\ \hline
$\sigma^2$ & 0.5 & 0.5 & 0.5 & 1 & 1 & 2 \\ \hline
\end{tabular}%
}
\end{table}

\begin{table}[t]
\caption{\textsc{MAP of CIFAR-10-Full. The results of DPSH$^*$ is reported in \cite{DSDH}. DPSH$^+$ denotes the results with GoogLeNet. Our method (DCWH) achieves the best performance}}
\label{table:CIFAR-10-Full}
\centering
\begin{minipage}[t]{0.4\textwidth}
\centering
\resizebox{\textwidth}{!}{%
\begin{tabular}{|c||c|c|c|c|}
\hline
\multicolumn{5}{|c|}{MAP of CIFAR-10-Full} \\ \hline
Method & 16 bits & 24 bits & 32 bits & 48 bits \\ \hline
DCWH & \textbf{0.940} & \textbf{0.950} & \textbf{0.954} & \textbf{0.952} \\ \hline
DSDH & 0.935 & 0.940 & 0.939 & 0.939 \\ \hline
DTSH & 0.915 & 0.923 & 0.925 & 0.926 \\ \hline
DPSH & 0.763 & 0.781 & 0.795 & 0.807 \\ \hline
DSRH & 0.608 & 0.611 & 0.617 & 0.618 \\ \hline
DPSH$^*$ & 0.903 & 0.885 & 0.915 & 0.911 \\ \hline
DPSH$^+$ & 0.905 & 0.902 & 0.923 & 0.905 \\ \hline
\end{tabular}%
}
\end{minipage}%
\end{table}

\begin{table}[t]
\caption{\textsc{MAP of CIFAR-10-Small. DSPH$^+$ and DSHNP are based on the GoogLeNet cited from \cite{DSHNP}. DCWH surpasses other state-of-the-art approaches with a clear margin}}
\label{table:CIFAR-10-Small}
\centering
\begin{minipage}{0.4\textwidth}%
\resizebox{\textwidth}{!}{%
\begin{tabular}{|c||c|c|c|c|}
\hline
\multicolumn{5}{|c|}{MAP of CIFAR-10-Small}                                          \\ \hline
Method   & 12 bits          & 24 bits          & 32 bits          & 48 bits          \\ \hline
DCWH     & \textbf{0.864} & \textbf{0.884} & \textbf{0.881} & \textbf{0.887} 		 \\ \hline
DSDH     & 0.740            & 0.786            & 0.801            & 0.820            \\ \hline
DPSH     & 0.713            & 0.727            & 0.744            & 0.757            \\ \hline
DTSH     & 0.710            & 0.750            & 0.765            & 0.774            \\ \hline
DQN      & 0.554            & 0.558            & 0.564            & 0.580            \\ \hline
DPSH$^+$ & 0.720            & 0.729            & 0.745            & 0.759            \\ \hline
DSHNP    & 0.731            & 0.758            & 0.766            & 0.784            \\ \hline
\end{tabular}%
}
\end{minipage}%
\end{table}

\begin{table}
\caption{\textsc{MAP of NUS-WIDE}}
\label{table:nus-wide}
\centering
\begin{minipage}{0.4\textwidth}
\resizebox{\textwidth}{!}{%
\begin{tabular}{|c||c|c|c|c|}
\cline{1-5}
  \multicolumn{5}{|c|}{MAP of NUS-WIDE} \\ \hline
  Method & 16 bits & 24 bit & 32 bit & 48 bit \\ \hline
  DCWH & \textbf{0.819} & \textbf{0.831} & \textbf{0.832} & \textbf{0.828} \\ \hline
  DSDH & 0.815 & 0.814 & 0.820 & 0821 \\ \hline
  DTSH & 0.756 & 0.776 & 0.785 & 0.799 \\ \hline
  DPSH & 0.715 & 0.722 & 0.736 & 0.741 \\ \hline
  DRSCH & 0.615 & 0.622 & 0.623 & 0.628 \\ \hline
  DSRH & 0.609 & 0.618 & 0.621 & 0.631 \\  \hline
\end{tabular}%
}
\end{minipage}
\end{table}

\begin{figure}[h]
\centering
\begin{minipage}{\textwidth}
\subfloat[Precision @Full]{\includegraphics[width=0.4\textwidth]{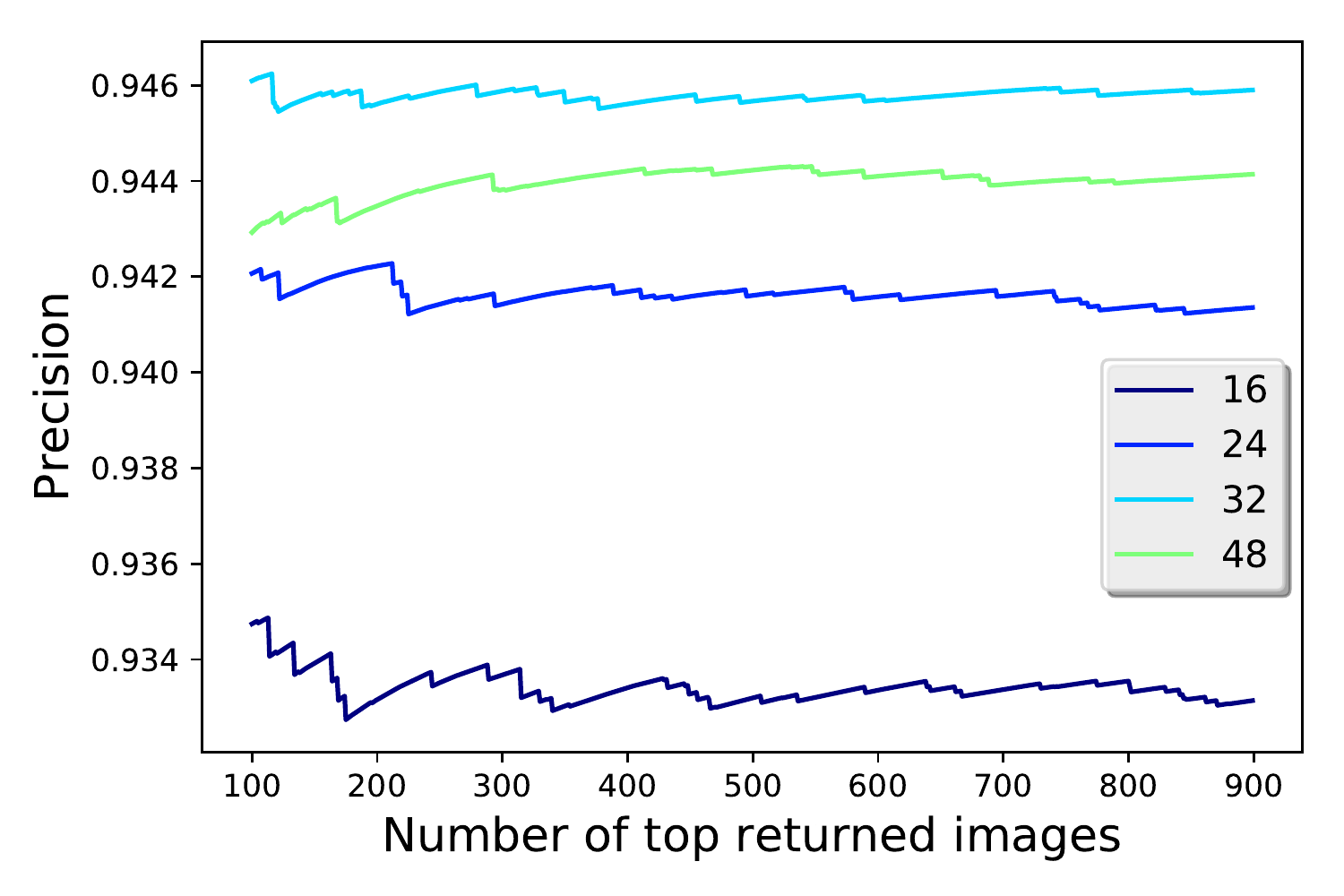}}
\\
\subfloat[Precision @Small]{\includegraphics[width=0.4\textwidth]{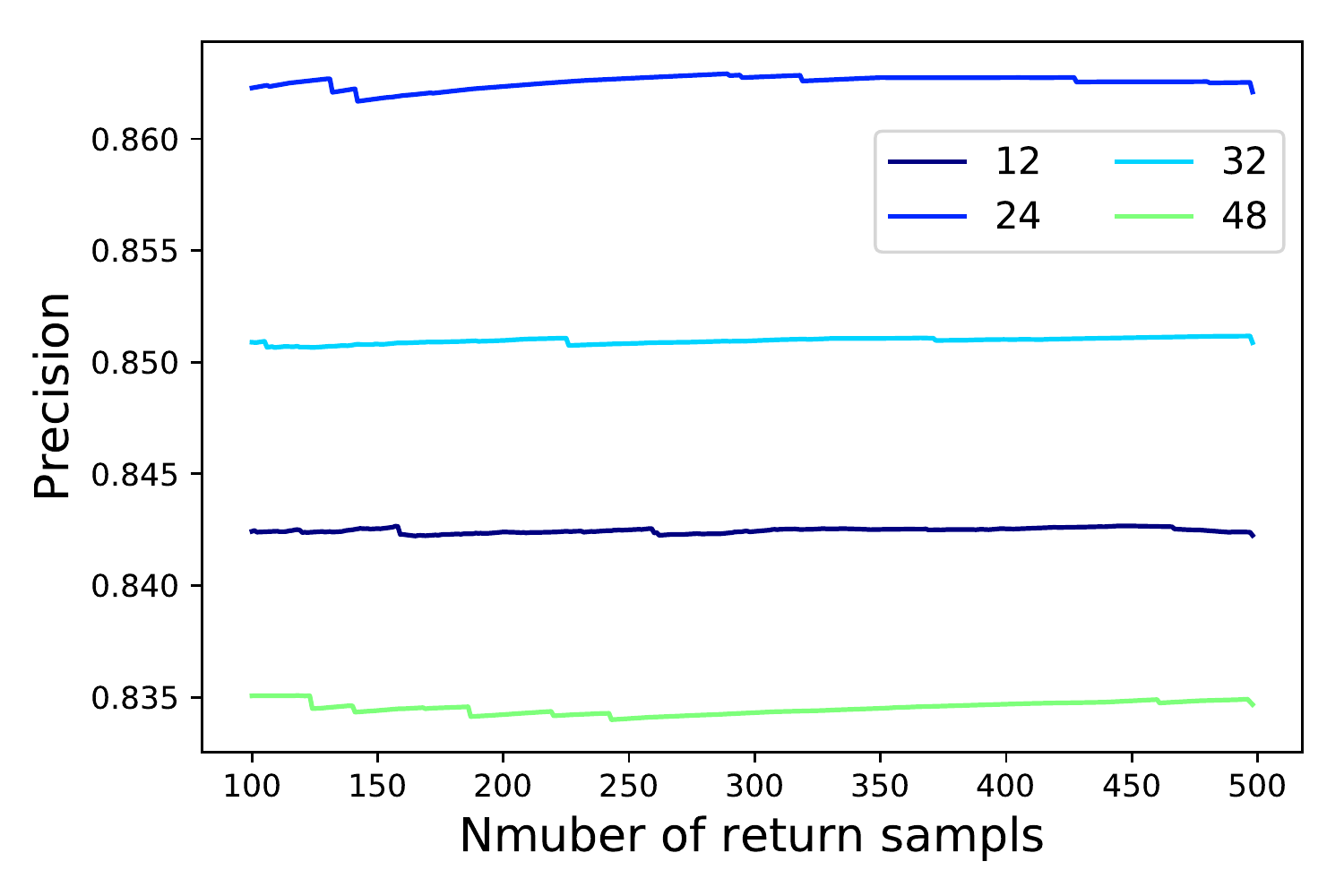}}
\end{minipage}
\caption{Precision of CIFAR-10.}
\label{fig:p_cifar}
\end{figure}

\subsection{Group I}

In this group, we follow the experiment setting in \cite{DPSH,DSDH}. 
Two important benchmark datasets: CIFAR-10 \cite{CIFAR10} and NUS-WIDE~\cite{chua2009nus} are used to evaluate the performance of several deep hashing methods.
The baseline deep hashing methods include DSDH~\cite{DSDH}, DSHNP~\cite{DSHNP}, DPSH~\cite{DPSH}, DQN~\cite{DQN}, DSRH~\cite{yao2016DSRH}, DTSH~\cite{DTSH} and CNNH~\cite{xia2014supervised}.

\begin{figure*}[t!]
	\centering
	\begin{minipage}{0.9\textwidth}
		\includegraphics[width=\textwidth]{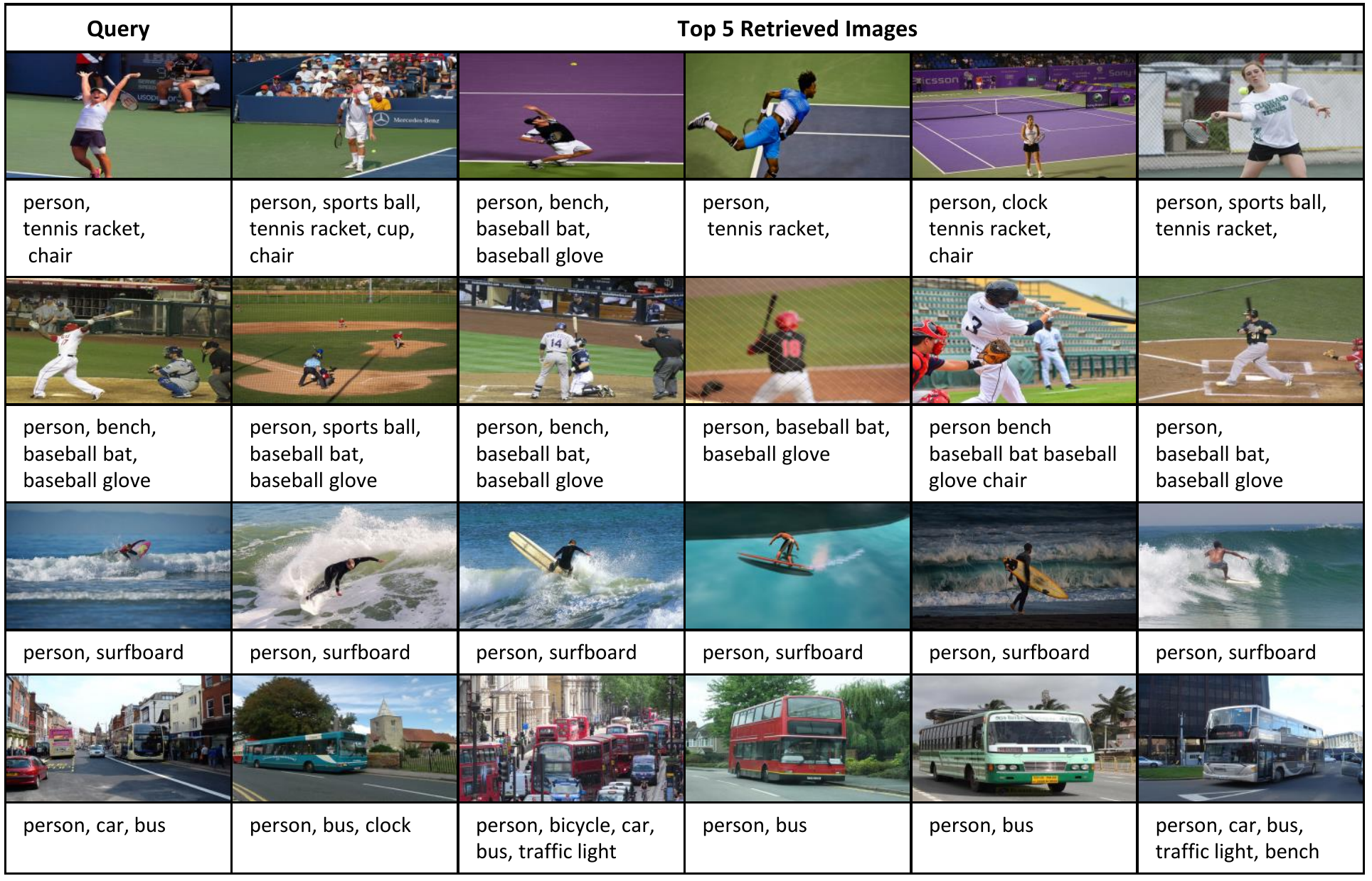}
		\captionof{figure}{Retrieval results of COCO @48-bit.}
		\label{fig:coco_results}
	\end{minipage}
\end{figure*}

\subsubsection{CIFAR-10}

CIFAR-10~\cite{CIFAR10} is a single label dataset, which is one of the most widely used datasets for evaluating image retrieval methods. 
It has 10 classes, and each class has $6,000$ color images with $32\times32$ resolution. 
Following \cite{DPSH,DSDH}, two experiment settings are considered in our experiment. 
For the first setting, we randomly take $1,000$ images per class ($10,000$ samples in total) as query images. The rest samples are used for training. 
We denote this experiment setting as CIFAR-10-Full. 
For the second, $500$ images per class are randomly selected as training set and $100$ images per class are randomly selected as query images. 
The second experiment setting is denoted as CIFAR-10-Small. 
The MAPs for CIFAR-10-Full and CIFAR-10-Small are evaluated on top $50,000$ and $5,000$ returned neighbors.

The MAP results of CIFAR-10-Full and CIFAR-10-Small are presented in Tables~\ref{table:CIFAR-10-Full} and~\ref{table:CIFAR-10-Small}, respectively. 
It can be found in Table~\ref{table:CIFAR-10-Full} that, when there are enough training samples, DCWH, DSDH, DPSH and DTSH can achieve more than $0.9$ MAP. 
Our method achieves the best performance $0.954$ at code length $32$, which is $1.4$ percents higher than the best performance achieved by DSDH at $24$ bits. 
For all bits, DCWH provides the highest performance among all methods. 
When training samples are reduced to $500$ images per class, the retrieval performance of several deep supervised hashing are dramatically decreased. 
By fully using the semantic label, the advantage of our method becomes more significant when training samples are limited. 
Shown in Table \ref{table:CIFAR-10-Small}, DCWH surpasses other deep hashing methods with a clear margin and achieves $0.887$ MAP at code length $48$.
For low bit cases, e.g., 12 bits, our method is significantly better than others deep hashing models. 
The average MAP of DCWH is $0.879$, which is $9$ percents higher than the average of DSDH's $ 0.787$. 
We further present the precision at different top returned samples in Figure~\ref{fig:p_cifar}. 
It shows that our method maintains a stable precision for both experiment settings.

\subsubsection{NUS-WIDE}
The NUW-WIDE \cite{chua2009nus} is a multi-label dataset, which collects around $270,000$ images from Flickr.com. 
Every image is annotated with one or several tags from 81 tags. 
Following \cite{DPSH,DSDH}, images labeled with 21 most frequent tags are used in this experiment.
Moreover, 100 images per tag are randomly selected as query images and the rest are used for training. 
The MAP is evaluated based on top $50,000$ neighbors and the results are presented in Table~\ref{table:nus-wide}. 
It can be seen from the table that DCWH achieves the state-of-the-art performance. 
The average MAPs of DCWH is $0.827$ and the average MAP of DSDH, DTSH and DPSH are $0.818$, $0.787$ and $0.733$ respectively. 
Our method is better than DSDH, which uses both the pair-wise and the class label. 
Compared with DPSH and DTSH, both of which only use the similarity label, DSDH and our model achieves higher performance. 
This indicates that fully using the semantic label can distinctly improve the semantic retrieval results. 

\begin{table}
\caption{\textsc{MAP of ImageNet-$100$}}
\label{table:ImageNet}
\centering
\begin{minipage}{0.4\textwidth}
\resizebox{\textwidth}{!}{%
\begin{tabular}{|c||c|c|c|c|}
\hline
\multicolumn{5}{|c|}{MAP of ImageNet-100}                                       \\ \hline
Method  & 16 bits         & 32 bits         & 48 bits         & 64 bits         \\ \hline
DCWH    & \textbf{0.7817} & \textbf{0.7987} & \textbf{0.8351} & \textbf{0.8490} \\ \hline
HashNet & 0.4420          & 0.6061          & 0.6633          & 0.6835          \\ \hline
DHN     & 0.3106          & 0.4717          & 0.5419          & 0.5732          \\ \hline
DNNH    & 0.2903         & 0.4605          & 0.5301          & 0.5645          \\ \hline
CNNH    & 0.2812          & 0.4498          & 0.5245          & 0.5538          \\ \hline
\end{tabular}%
}
\end{minipage}
\end{table}

\begin{table}
\caption{\textsc{MAP of COCO}}
\label{tab:coco}
\centering
\begin{minipage}{0.4\textwidth}
\resizebox{\textwidth}{!}{%
\begin{tabular}{|c||c|c|c|c|}
\hline
\multicolumn{5}{|c|}{MAP of COCO}                                               \\ \hline
Method  & 16 bits         & 32 bits         & 48 bits         & 64 bits         \\ \hline
DCWH    & \textbf{0.7418} & \textbf{0.7762} & \textbf{0.7856} & \textbf{0.7789} \\ \hline
HashNet & 0.6873          & 0.7184          & 0.7301          & 0.7362          \\ \hline
DHN     & 0.6774          & 0.7013          & 0.6948          & 0.6944          \\ \hline
DNNH    & 0.5932          & 0.6034          & 0.6045          & 0.6099          \\ \hline
CNNH    & 0.5642          & 0.5744          & 0.5711          & 0.5671          \\ \hline
\end{tabular}%
}
\end{minipage}
\end{table}
\subsection{Group II}
In the second group of experiments, we follow the experiment setting in HashNet~\cite{2017hashNet}. 
The main challenge of this group is that only part of images in database are used for training the hashing function. 
This is closer to real application. 
We evaluate the performance on other two benchmark image retrieval datasets: ImageNet~\cite{ILSVRC15} and MS COCO~\cite{coco}. 
The results of HashNet, DHN~\cite{zhu2016DHN}, CNNH~\cite{xia2014supervised} and DNNH~\cite{lai2015DNNH} are directly taken from~\cite{2017hashNet}. 
For both experiments, we use the same image partitions as the HashNet~\cite{2017hashNet}\footnote{{https://github.com/thuml/HashNet}}.

\textbf{ImageNet}~\cite{ILSVRC15} is a single-label image dataset for Large Scale Visual Recognition Challenge (ILSVRC 2015). 
It contains $1,000$ categories and each category has around $1,200$ images in training sets.
As suggested in~\cite{2017hashNet}, $100$ categories are randomly selected and all the images in the training set of these categories are used as the retrieval database. 
All the images in validation set are used as query images. 
Moreover 100 images per class are used for training. The main challenge of this experiment setting is that the training samples for each class are very limited (100 images per class). 
The MAP is evaluated based on top $1,000$ retrieved images.

\textbf{MS COCO}~\cite{coco} is a large-scale object detection, segmentation, and captioning dataset. 
Each image is assigned with some of $8$0 categories.
$5,000$ images are randomly selected from the whole dataset as query images. 
The rest images are used as the database and $1,0000$ images are randomly selected as training points. 
The MAP is adopted on top $5,000$ returned images for evaluation.

The MAP results on ImageNet and COCO are presented in Tables~\ref{table:ImageNet} and~\ref{tab:coco}. 
For the ImageNet-100, our method achieves the best MAP, $0.8490$ at $64$ bits, which is $16.5$ percents better than the best performance achieved by HashNet at the same bits. 
At $16$ bits, DCWH surpasses HashNet more than $30$ percents. 
The average MAP of our model is $0.8162$, which is $21.7$ percents higher than the mean average of HashNet ($0.5987$). 
It clearly shows that our model is better to learn a compact binary representation compared with other state-of-the-art methods. 
At code length $48$ bits and $64$ bits, our model is slightly better than HashNet. 
As for the COCO dataset, DCWH surpasses the HashNet by $4$ to $5$ percents, and the best performance, $0.7856$ is achieved at $48$ bits. 
Figure~\ref{fig:coco_results} shows four retrieval examples of our model at code length $48$ bits.
Four query images share one label of person and other labels are different. 
If only similar/dissimilar labels are considered, four query images are similar to each other. 
With the help of semantic labels, our system returns more relevant images. 
For example, images denoted with person are correct retrieval results for the first query image. 
By fully utilizing the class label information, our system not only returns images of person, but also retrievals images of person playing tennis, which are more user-desired results in terms of semantic retrieval.  

\section{Discussion}
\label{Sec:discussion}
\subsection{Performance under NDCG metric}
Besides the mean average precision (MAP), Normalized Discounted Cumulative Gain (nDCG)~\cite{jarvelin2002cumulated} is another important metric to evaluate the retrieval performance. 
We follow the experiment setting in~\cite{wang2017Hierarchical} and verified performance on the CIFAR-100~\cite{CIFAR10}. 
We randomly selected $90\%$ images as training set and the remaining $10\%$ as the query set. The nDCG results at top 100 returned images are shown in Table~\ref{tab:ndcg}. 
From this table, it can be found that our method surpasses SHDH and DPSH with a distinct margin and achieves best performance with $0.7547$ in terms of nDCG, which is  higher more than $11.41$ percents compared with previous best one. 
\begin{table}[h]
\caption{nDCG @100 of CIFAR-100}
\label{tab:ndcg}
\centering
\resizebox{0.33\textwidth}{!}{%
\begin{tabular}{|c||c|c|c|}
\hline
\multicolumn{4}{|c|}{nDCG @100 of CIFAR-100}                  \\ \hline
Method & 32 bits         & 48 bits         & 64 bits         \\ \hline
DCWH   & \textbf{0.7546} & \textbf{0.7547} & \textbf{0.7506} \\ \hline
SHDH\cite{wang2017Hierarchical}   & 0.6141          & 0.6281          & 0.6406          \\ \hline
DPSH   & 0.5650          & 0.5693          & 0.5751          \\ \hline
\end{tabular}%
}
\end{table}
\subsection{Performance on Different Stages}

\begin{figure}[h]
    \centering
    \includegraphics[width=0.4\textwidth]{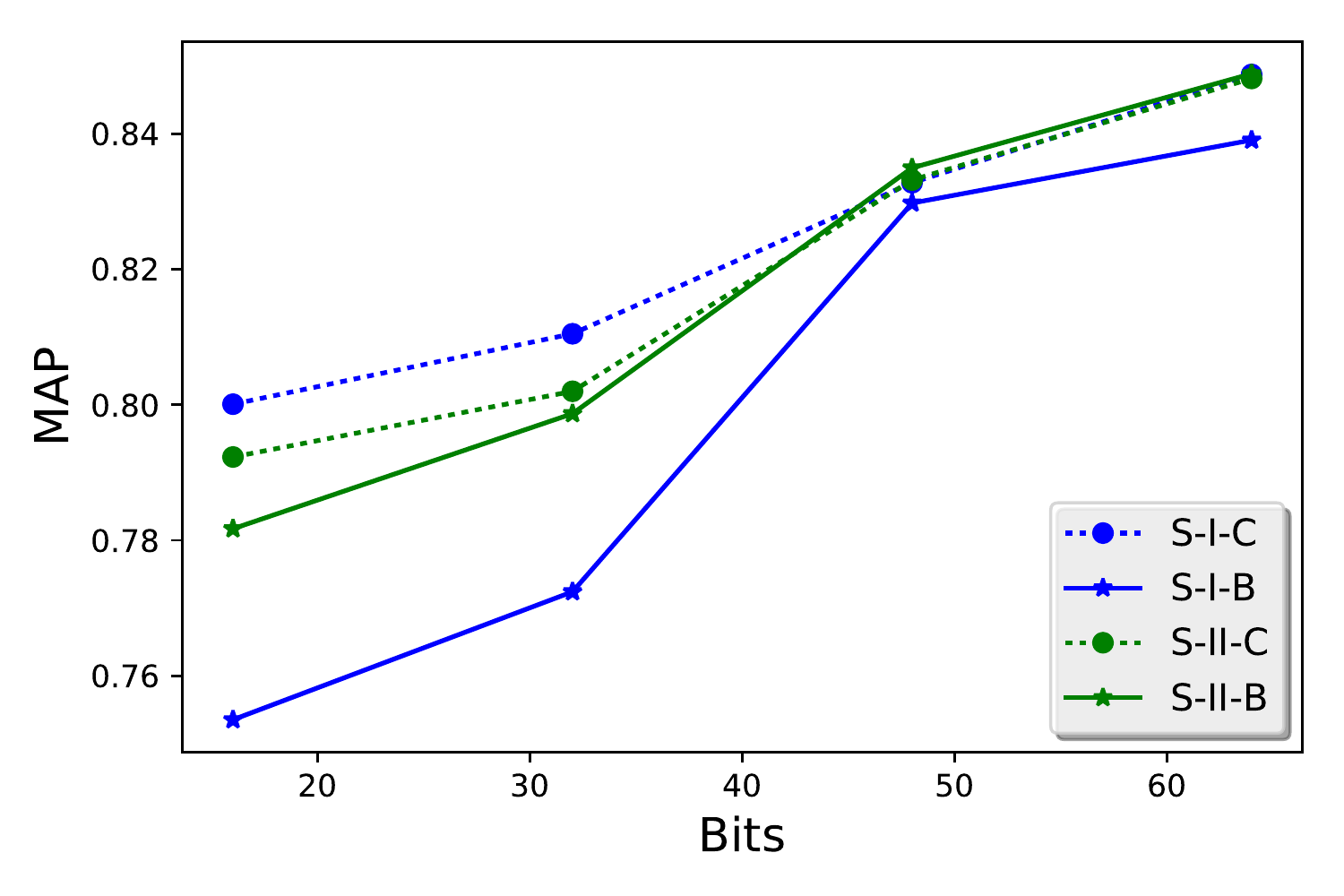}
    \caption{Performance on different stages. The second stage narrows the performance gaps between binary codes and continous features. When the code lengths is large enough, the first stage can achieve the close performance to the second. }
    \label{fig:stages}
\end{figure}

To better understand the proposed two-stage optimization strategy, we evaluate the retrieval results of binary codes and continuous features on \textbf{Stage I} and \textbf{Stage II}. 
ImageNet-100 is used in this experiment as an example dataset. 
The MAP of the binary codes are denoted as \textbf{I-B} and \textbf{II-B}, the retrieval results of continuous features (without the $sign$ function) are assigned with \textbf{I-C} and \textbf{II-C}. %
The results are presented in Table~\ref{tab:Stage}. 
By comparing the MAP of binary codes, we find that Stage II helps to reduce the quantization error as designed.
\begin{table}[h]
\caption{\textsc{Performance on different stages. \textbf{C} indicates the retrieval results of continuous features (\textbf{without} $sign$). \textbf{B} represents the results of binary codes (\textbf{with} $sign$)}}
\label{tab:Stage}
    \centering
    \resizebox{0.4\textwidth}{!}{%
\begin{tabular}{|c||c|c|c|c|}
\hline
\multicolumn{5}{|c|}{MAP of ImageNet-100}         \\ \hline
stage     & 16 bits & 32 bits & 48 bits & 64 bits \\ \hline
I-C & 0.8001  & 0.8105  & 0.8382  & 0.8488  \\ \hline
I-B & 0.7535  & 0.7724  & 0.8289  & 0.8391  \\ \hline
II-C & 0.7923  & 0.8020  & 0.8332  & 0.8482  \\ \hline
II-B & 0.7817  & 0.7987  & 0.8350  & 0.8490  \\ \hline
\end{tabular}%
}
\end{table}
We further plot MAP results in Figure~\ref{fig:stages} to show the performance on different bits.
The best performance given by the continuous features is in Stage I. 
After refining the DCWH model with the quantization error (Stage II), the performance with binary codes increases but the performance of continuous features slightly decrease. 
The difference between continuous and binary codes in Stage II is much smaller than that in Stage I.
In Stage I, it can be observed that the gap between the MAP of continuous features and the MAP of binary codes becomes smaller when the dimension (the number of bits) increases. 
Part of the reason is that when the dimension is higher, the ``corner part'' of the hypercube takes over of the most of the volume of the hypercube. 
The projected points have a higher probability falling into the corner part where the quantization error is small.
\subsection{Update Class Centers by Gradient Descent}
\begin{figure}[h]
	\centering
	\subfloat[Training curves of DCWH-G]{\includegraphics[width=0.25\textwidth]{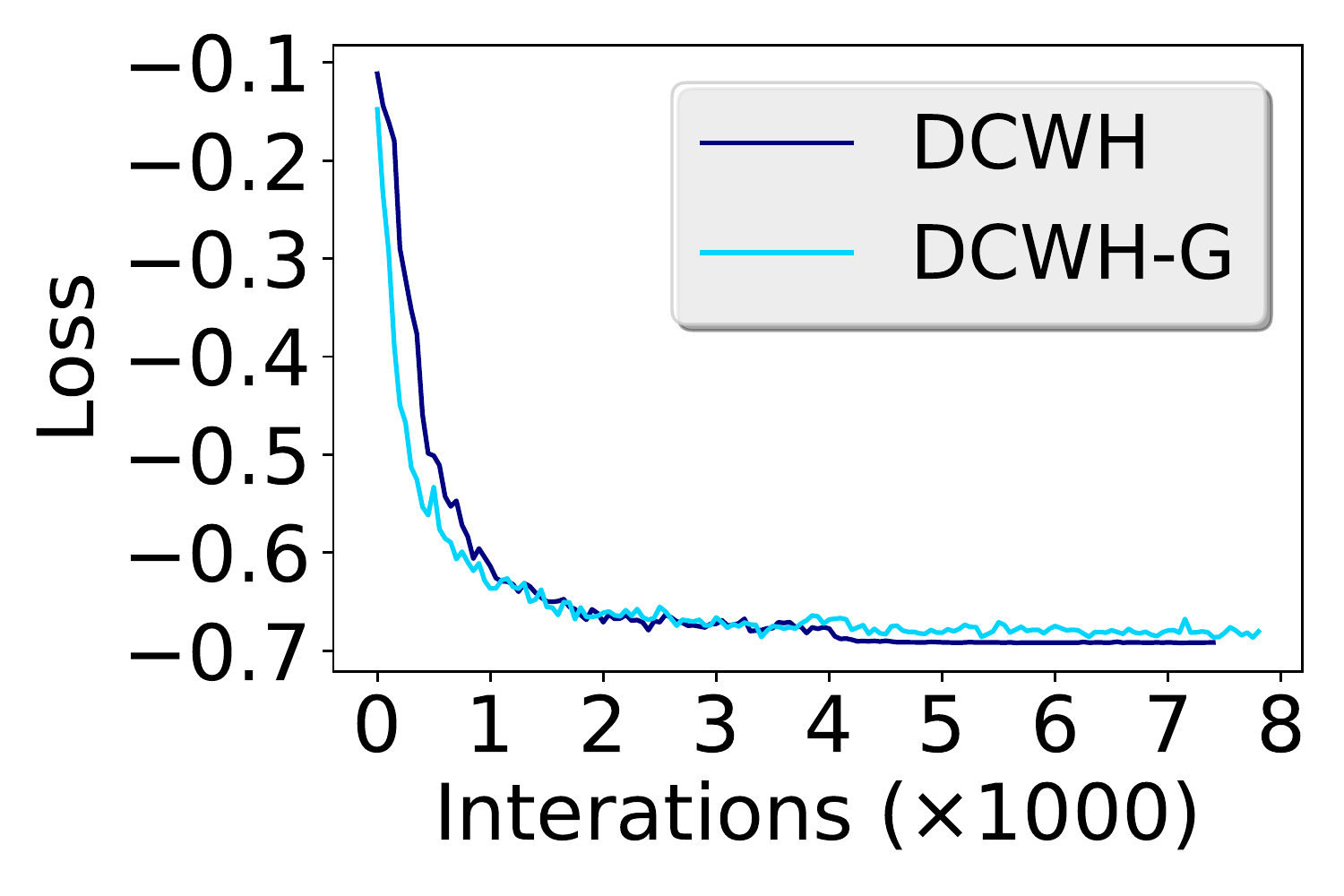}\label{fig:lossDCWH_g}}
	\subfloat[MAP of CIFAR-10-Small]{\includegraphics[width=0.25\textwidth]{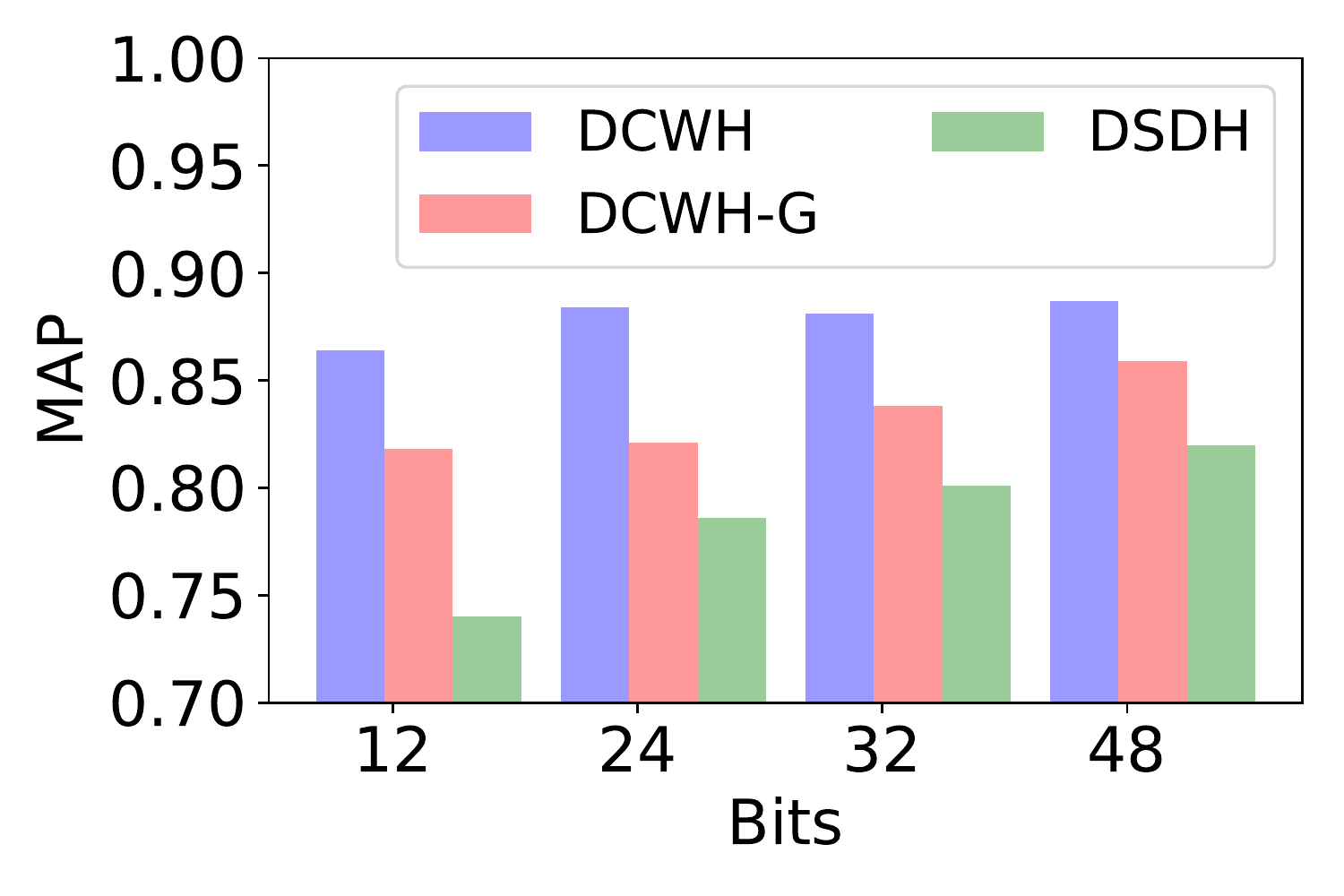}\label{fig:map_DCWH_G}}
	\caption{Results of Gradient-updating. DCWH-G indicates the results of gradient updating class centers. It shows that DCWH-G can achieve the state-of-the-art performance but not as good as DCWH.\label{fig:DCHW-G}}
\end{figure}

Instead of periodically updating the class centers $\{\mu_i\}$ as shown in Algorithm~\ref{alg:1}, the class centers also can be treated as parameters of the CNN and updated every iteration with the gradient. 
The derivative of $\{\mu_i\}$ can be easily obtained from our objective function. 
Following ~\cite{Center_lossWen2016,liu2017coco_loss}, the class centers $\{\mu\}$ are firstly  initialized by calculating class centers of training data then updated by back-propagation. 
We verified this idea on CIFAR-10-small and present the results in Figure~\ref{fig:DCHW-G}\subref{fig:map_DCWH_G} and the training curve in Figure~\ref{fig:DCHW-G}\subref{fig:lossDCWH_g}. 
Though DCWH-G produces slightly lower MAP, it has a close convergence speed as DCWH and does not need to periodically forward training set to update class centers. 
One possible reason why DCWH-G provides lower MAP is that the class centers updated by gradient are not as accurate as periodical-updating ones. 
We leave the improvement as future work.
\subsection{Impact of $\sigma^2$  }
\begin{figure}[h]
    \centering
    \subfloat[Intra-class variance]{\includegraphics[width=0.4\textwidth]{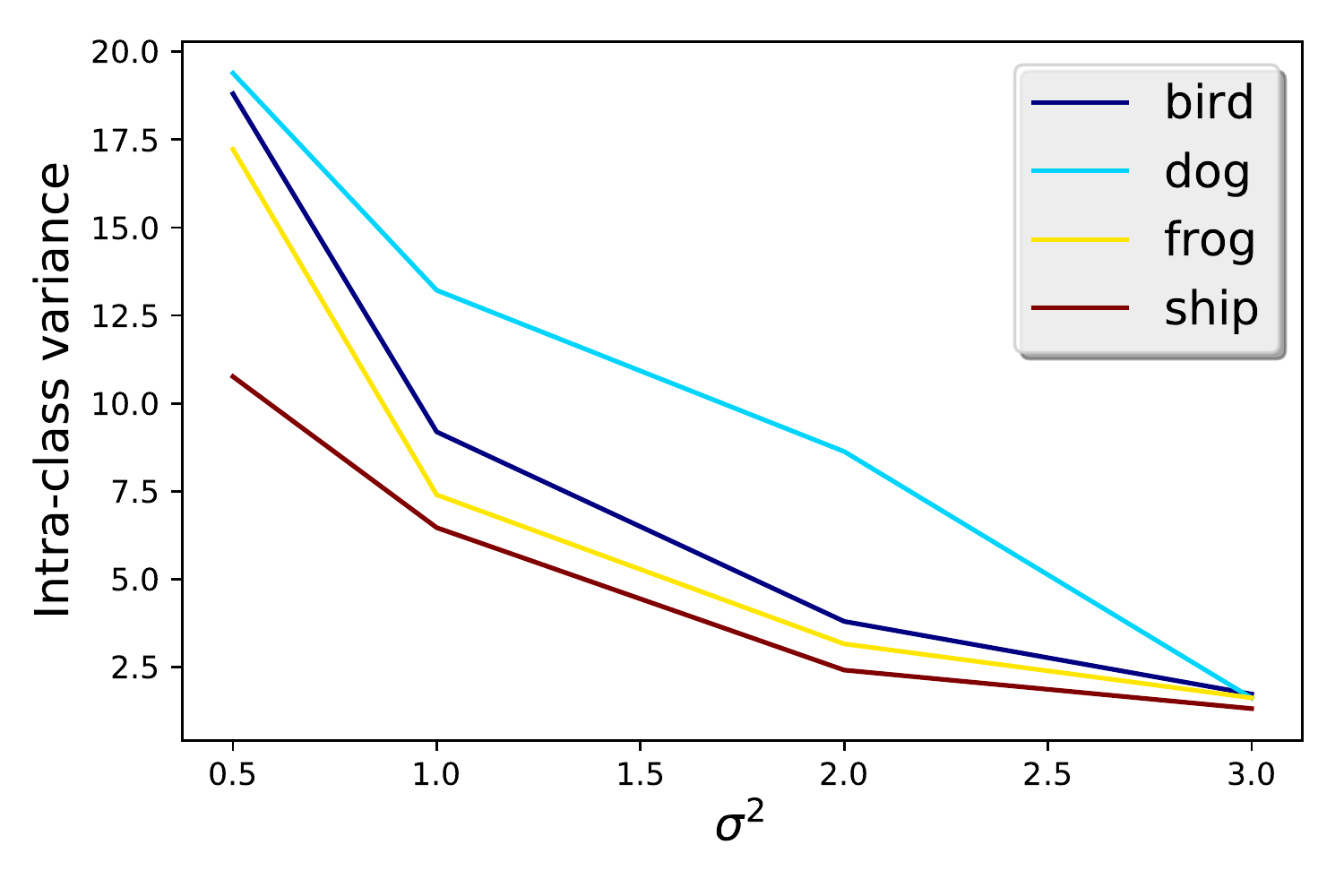}\label{fig:variance}}\\
    \subfloat[Average inter-class distance ]{\includegraphics[width=0.4\textwidth]{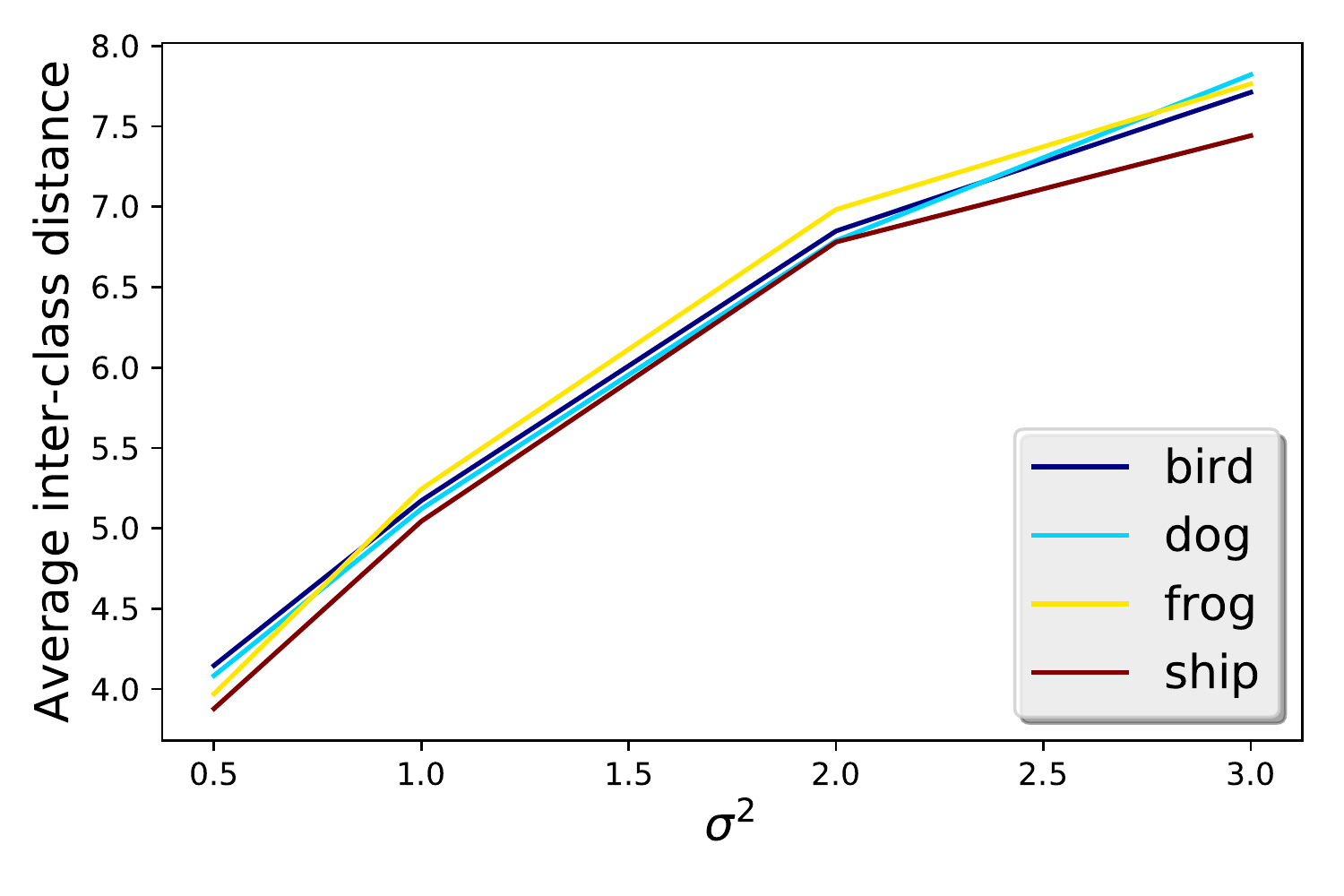}\label{fig:interclass}}\\
\caption{Intra-class variance and average inter-class distance at 32 bits.\label{fig:sigma}}
\end{figure}

In this section, we experimentally investigate how the hyper parameters $\sigma$ influences our model. 
The CIFAR-10-small is chosen as the example dataset to conduct the following experiment. 
We train DCWH with different combinations of bits and $\sigma^2$. The MAP results are presented in Table~\ref{tab:sigma}. 
From the MAP results, it can be observed that 12 bits and 24 bits work best with $\sigma^2=0.5$ and for 32 bits and 48 bits, the best choice of $\sigma^2$ is $1$. 
Setting a large $\sigma^2$ for a small feature space can lead to an extremely low MAP. 
For example, the MAP of $12$ bits with $\sigma^2=3$ is lower than $0.20$.
To explore how $\sigma^2$ influence the class distributions in the feature space, we fix the code length to $32$ bits and calculate the intra-class variances and average inter class distances of different $\sigma^2$ based on the training data. 
We randomly choose the four classes and present the intra-class variances change in Figure~\ref{fig:sigma}\subref{fig:variance} and the average inter class distances in Figure~\ref{fig:sigma}\subref{fig:interclass}. 
It can be observed that with the increasing of $\sigma^2$, the intra-class variances of all four classes significantly reduce. Meanwhile, the average inter-class distances are increased.
Given the code length and number of classes, the selection of $\sigma^2$ is a trade-off between the inter-class gap and the intra-class divergence. 
A small value of $\sigma^2$ leads to a large intra-class variance but a vague inter class gap. 
On the contrary, slightly large values of the $\sigma^2$ cause a small intra-class variance, which makes the feature space hard to capture the intra-class divergence. 
In general, the low dimension works fine with small $\sigma^2$ and choosing slightly larger $\sigma^2$ for higher dimension can produce better results.  
\begin{table}[h]
\caption{\textsc{MAP on different $\sigma^2$}}
\label{tab:sigma}
\centering
\resizebox{0.4\textwidth}{!}{%
\begin{tabular}{|c||c|c|c|c|}
\hline
\multirow{2}{*}{$\sigma^2$} & \multicolumn{4}{c|}{CIFAR-10-Small} \\ \cline{2-5} 
 & 12 bits & 24 bits & 32 bits & 48 bits \\ \hline
0.5      & 0.864   & 0.884   & 0.839   & 0.878   \\ \hline
1        & 0.831   & 0.872   & 0.881   & 0.887       \\ \hline
2        & 0.331   & 0.838   & 0.865   & 0.867     \\ \hline
3        & 0.182   & 0.812   & 0.845   & 0.857   \\ \hline
\end{tabular}%
}
\end{table}
\subsection{Convergence Speed}
\begin{figure}[h]
    \centering
    \subfloat[Different bits] {\includegraphics[width=0.23\textwidth]{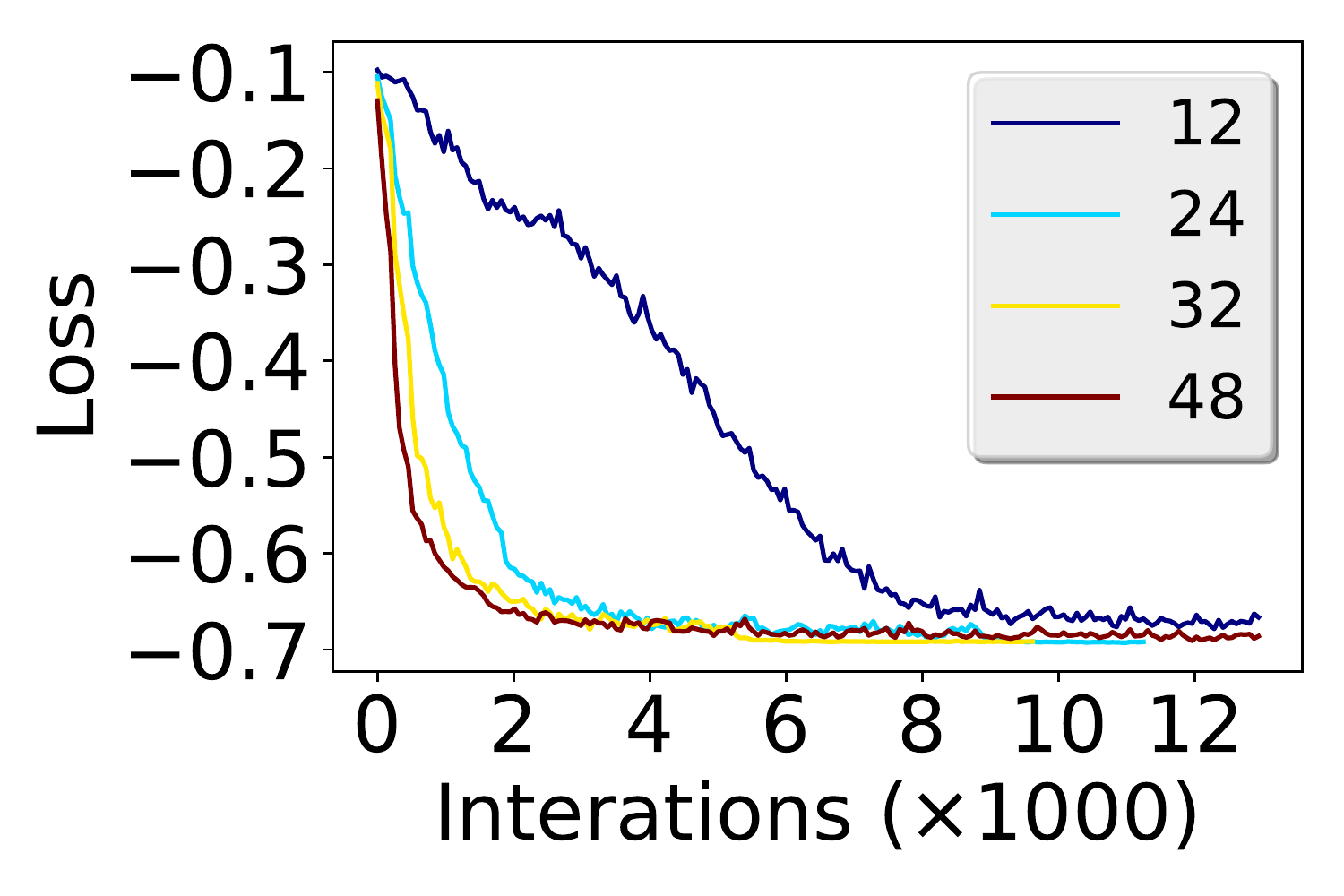}\label{fig:loss_bits}}\hfil
    \subfloat[Different $\sigma^2$] {\includegraphics[width=0.23\textwidth]{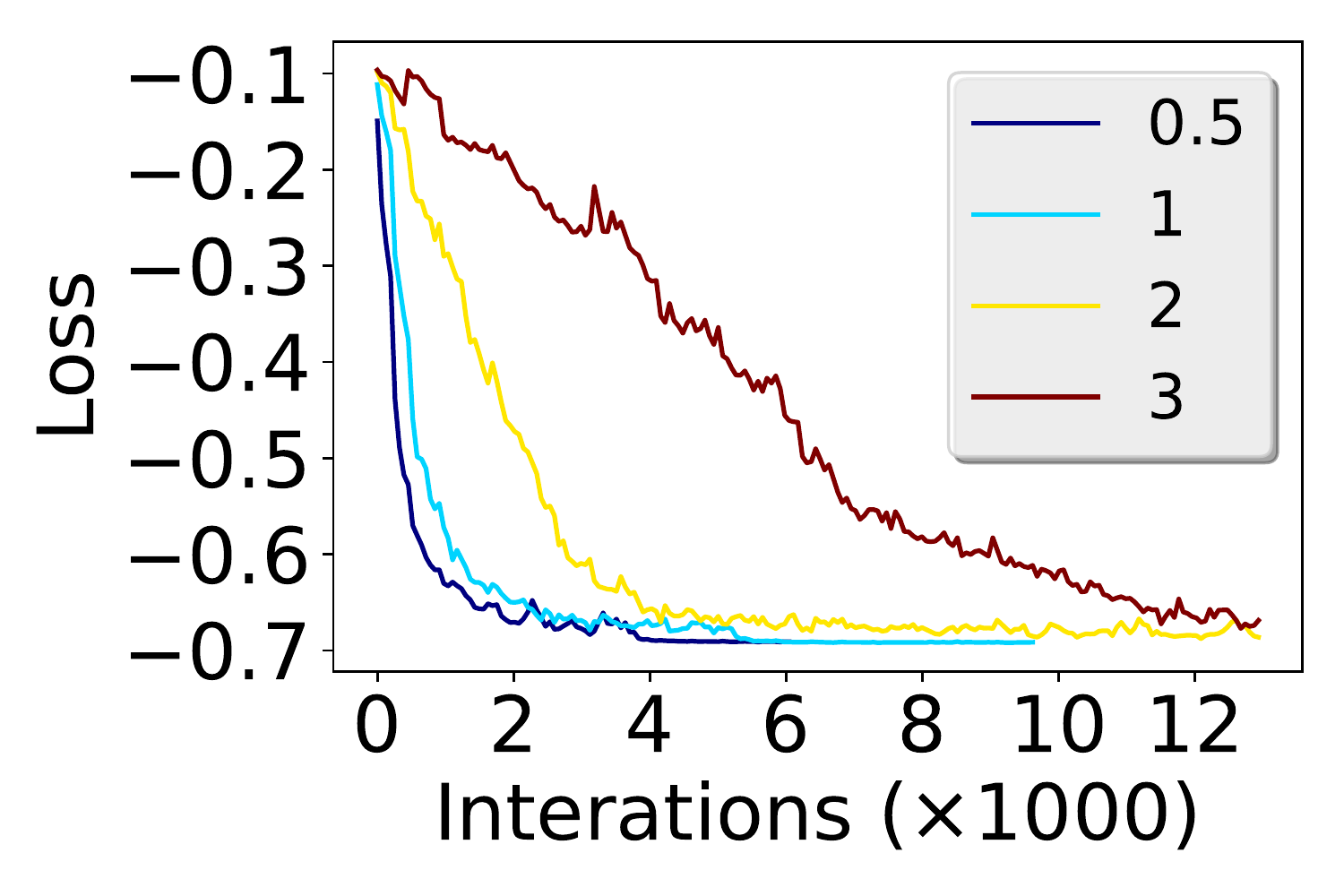}\label{fig:loss_sigma}}\\
    \caption{Training curves of CIFAR-10-Small.\label{fig:loss}}
\end{figure}

The first stage of our optimization strategy takes over the most training time. 
In this section, we show how the length of codes and the value of $\sigma^2$ influence the convergence speed of the first stage. 
We train our model with different bit lengths and $\sigma^2$ on the CIFAR-10-small. 
First we fix the $\sigma^2 = 1$, and plot the training curve of different bits in Figure~\ref{fig:loss}\subref{fig:loss_bits}. 
It can be observed that the model with a long code length converges faster than that with the shorter one. 
Then we fix the code length to 32 bits, and test with different $\sigma^2$. 
The training curves of different bits are shown in Figure~\ref{fig:loss}\subref{fig:loss_sigma}. 
It clearly shows that model with large value $\sigma^2$ converges slower than the small value of $\sigma^2$.

\begin{figure}[h]
\centering
\begin{minipage}{0.45\textwidth}
\centering
\includegraphics[width=0.8\textwidth]{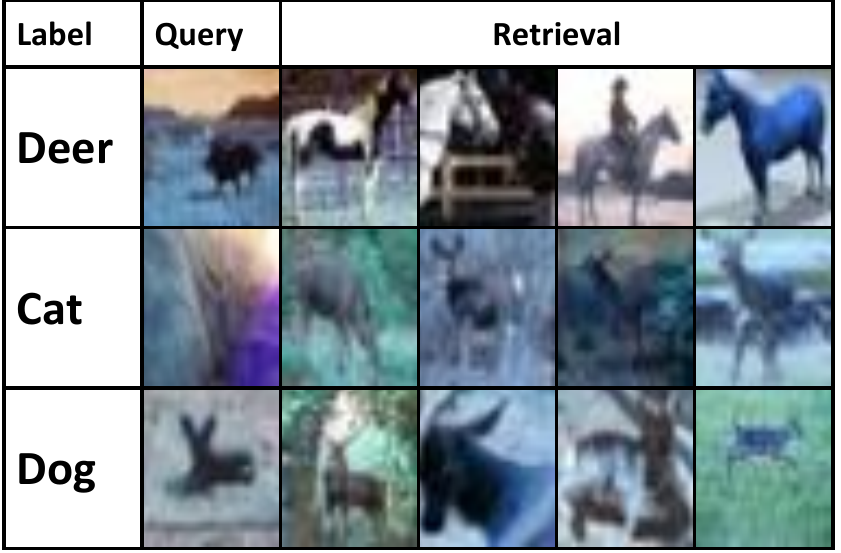}
\end{minipage}
\caption{Hard examples of CIFAR-10-Full. ``Label'' indicates the class labels of query images.}
\label{fig:hard_cifar}
\end{figure}

\subsection{Visualization of Hashing Codes}
\begin{figure}[h]
\centering
\begin{minipage}{0.48\textwidth}
\centering
\subfloat[ CIFAR-10-Full]{\includegraphics[width=0.45\textwidth]{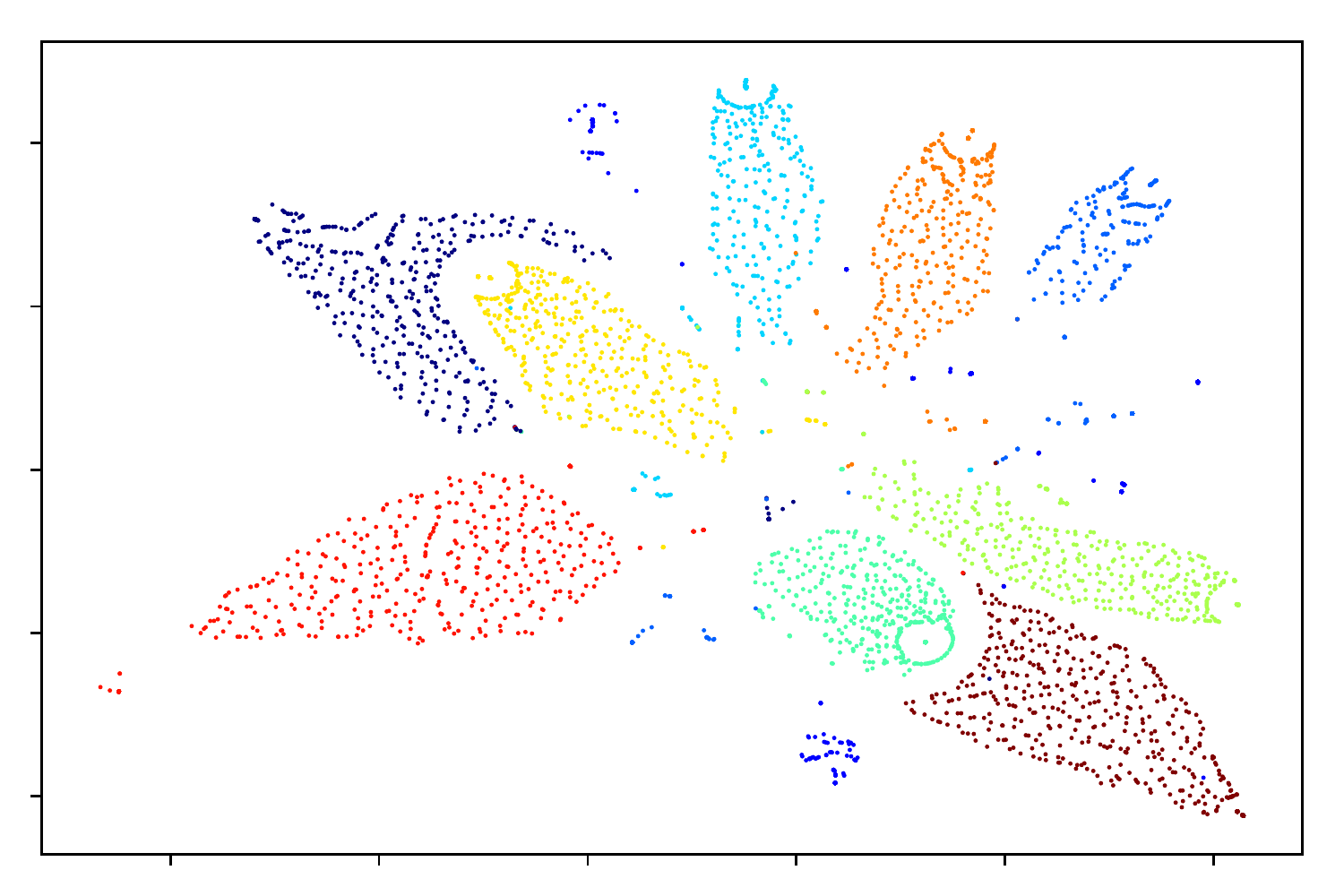}\label{fig:t-sne1}}
\subfloat[ImageNet-100]{\includegraphics[width=0.45\textwidth]{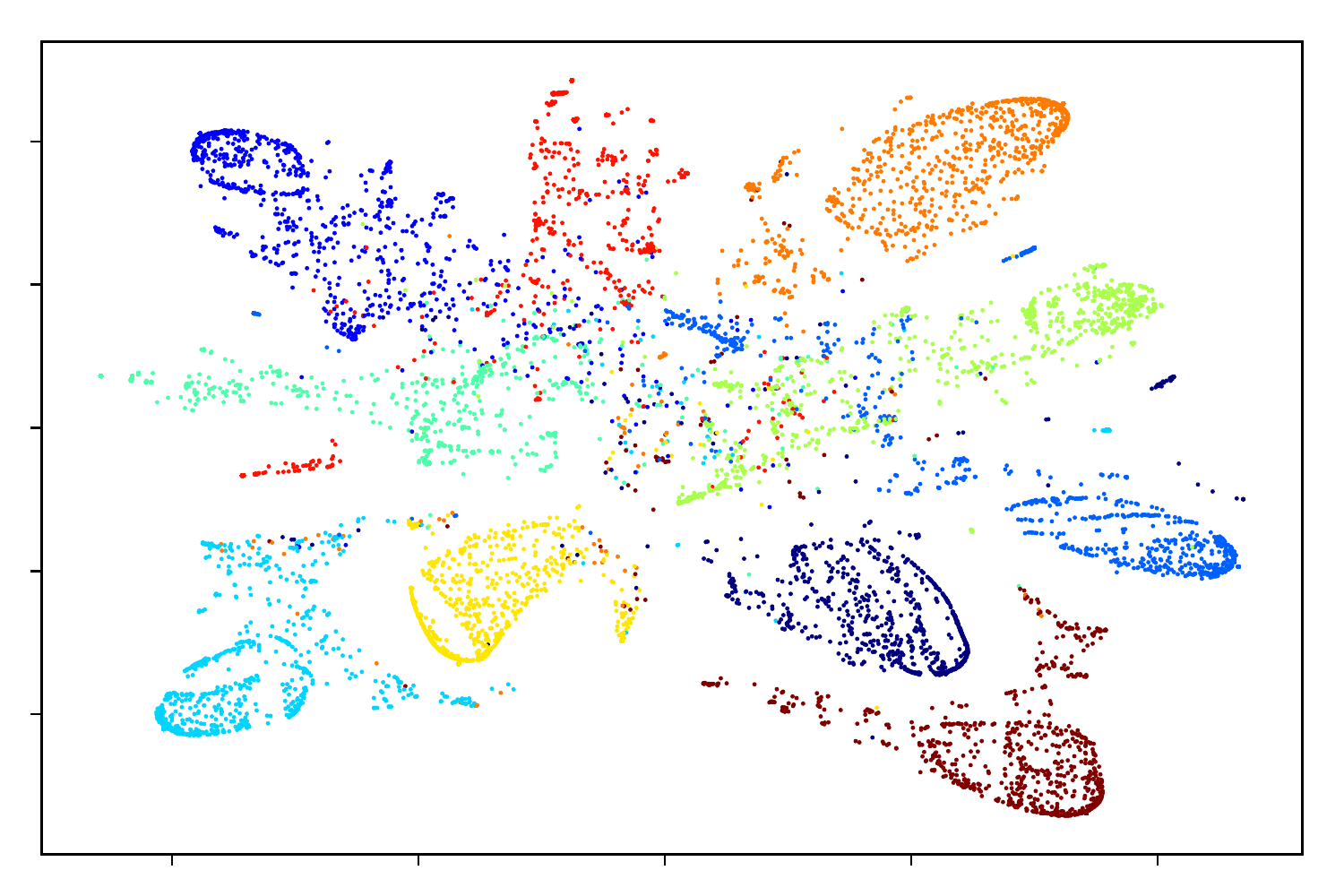}\label{fig:t-sne2}}
\end{minipage}
\caption{t-SNE visualization of hashing codes. Different colors indicate points from different classes. In (b), 10 classes are randomly selected from the  ImageNet-100 data set used in the experiments.  \textbf{Better viewed in color.}\label{fig:tsne}}
\end{figure}

To better illustrate the quality of hashing codes generated by DCWH, we use t-SNE \cite{maaten2008t-SNE} to visualize hashing codes in a 2D space. 
For CIFAR-10, we select 500 images of each class from the training set and generate 32-bits hash codes with the model in the experiment on CIFAR-10-full. 
The results are shown in Figure~\ref{fig:tsne}\subref{fig:t-sne1}. 
As for ImageNet, we only selected 10 categories from 100 categories used in the ImageNet-100, and select 500 images of each class. The visualization is presented in Figure~\ref{fig:tsne}\subref{fig:t-sne2}
From both figures, we can find that DCWH learns a discriminant structure in feature space. 
In this learnt Hamming space, most of points from the same class can be gathered together and different classes have relatively obvious inter-class gaps. 
With such a discriminant feature structure, retrieval can be completed efficiently and precisely.
\begin{figure*}[h!]
	\centering
	\begin{minipage}{0.9\textwidth}
		\includegraphics[width=\textwidth]{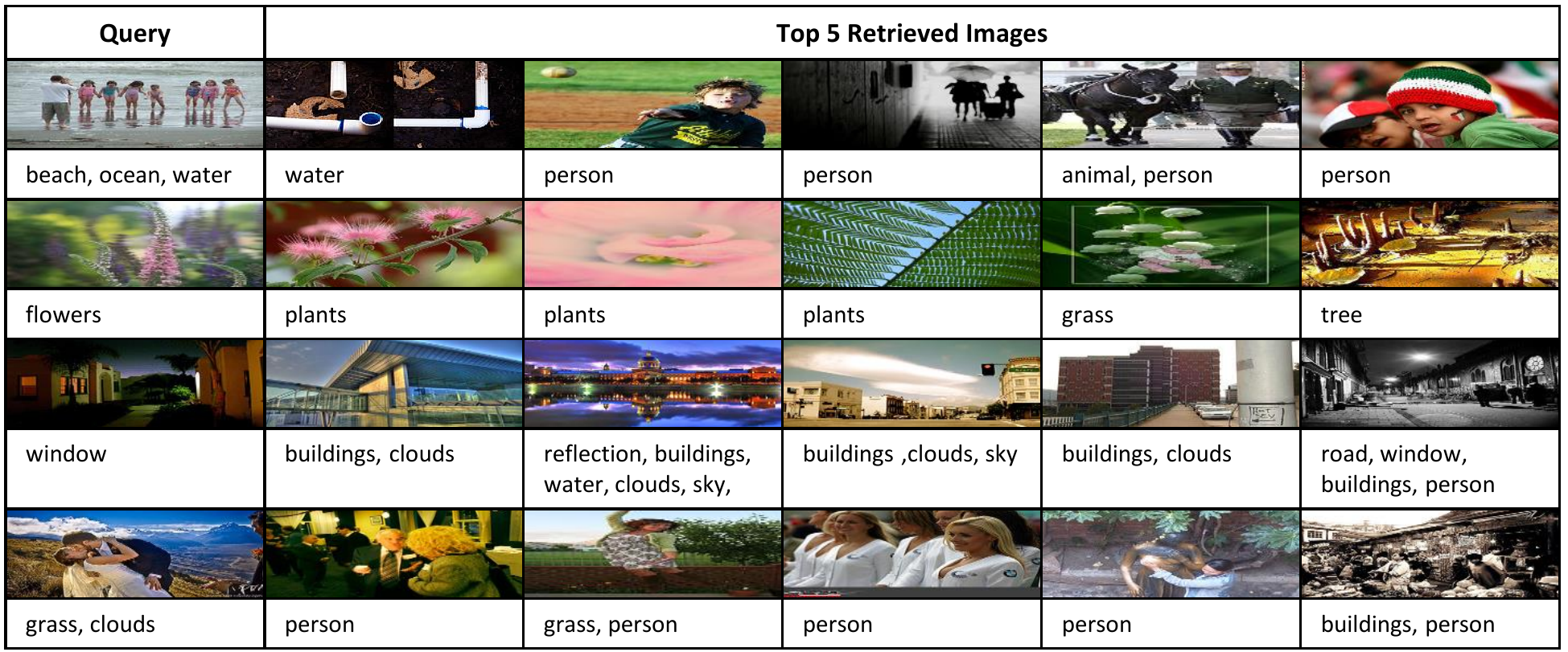}
	\end{minipage}
	\caption{Hard examples of NUS-WIDE.}
	\label{fig:hard_nus}
\end{figure*}
\subsection{Hard Example Analysis}
In this section, we randomly collect some examples with lower average precision from experiment group I to analyze how our system is failed. 
From Figure~\ref{fig:hard_cifar}, it can be seen that the main challenge of the CIFAR-10 dataset is the tinny size of target objects and low quality of images. 
The original size of images is $32\times32$, the objects of some image are extremely small and most detail of object is lost, which makes it hard to recognize the objects even for humans. 
Taking the third query image as an example, the original label of image is ``dog''. Our system retrieves deer images. 
This result is not correct but reasonable. 
Because only the contour of object can be seen, and the head part in the image looks very similar to deer's. 
For the second case, it is even hard to find where the cat is. 
To understand the performance on the NUS-WIDE, we analyze hundreds of query results and present four hard examples in Figure~\ref{fig:hard_nus}. 
One challenge of the NUS-WIDE is the problem of incomplete labels. The first and the fourth queries both miss the ``person'' label, which are the salience parts of images. 
For both images, our system returns the images of person. 
Another challenge is label overlapping. 
The second query is denoted with ``flower''. 
The top five retrial images include ``plants'', ``grass'', ``tree''. 
It can be found that ``plants'' should include ``flowers'', ``grass'' and ``tree''. 
There are other labels sharing the same problem, such as ``lake'', ``ocean'' and ``water''.

\section{Conclusion and Future Works}

In this paper, we propose a novel end-to-end deep hashing model, named as Deep Class-Wise Hashing (DCWH). 
More specifically, we first propose a class-wise objective function to fully using class label information. 
Then a two-stage optimization strategy is developed to effectively solve the discrete optimization problem introduced by binary codes. 
For every stage, an alternative learning process is proposed to train the feature learning part.
Extensive experiments show that our method can surpass other state-of-the-art ones in terms of image retrieval. 

For the future work, we will provide a more analytic way to decide hyper parameters (e.g., $\{\sigma_i\}$) and conduct further tests on large-scale data sets. 
As the proposed method can effectively learn a discriminant feature space, developing a more generic model for deep representation learning will be an important part of future work. 
Besides the class labels, the similar/dissimilar labels are also an important type of supervised information for visual data. 
Though the desired supervised information for our model is class-labels, it can be extended to similar/dissimilar data. 
Give an anchor image, $K$ similar images and $H$ dissimilar images are randomly selected to form a mini-batch. 
The anchor image and $K$ similar images are from the same class, and the class center of the anchor image can be calculated by $\hat{\mu}_{a} = \frac{1}{K} \sum_{i \in K} {r}_i$. 
The $H$ images are treated as $H$ different classes centers. 
The objective function is changed to
  
\begin{IEEEeqnarray}{ll}
&\mathrm{J}= - \sum_{a=1}^{N} \log \\
&\frac{\exp(-\frac{1}{2 \sigma^2} D^2(r_{a},\hat{\mu}_{a}))}   
{\sum_{i\in H} \exp(-\frac{1}{2 \sigma^2} D^2(r_{a},r_i)) +\exp(-\frac{1}{2 \sigma^2} D^2(r_a,\hat{\mu}_{a}))},\nonumber
\end{IEEEeqnarray}
where $r_n \in \{-1,1\}^L$ are binary codes with $L$ bits. 
Using similar/dissimilar pairs does not need to update the class center, but extra processes are needed to select images to generate mini-batches with the specific structure. 
This extension share close idea with N-pair deep metric learning~\cite{n-pairs}. 
We will investigate this extension in the future. 
\section*{Acknowledgments}
This work is supported by the Hong Kong Research Grants Council (Project C1007-15G), City University of Hong Kong (Project 9610034), and Shenzhen Science and Technology Innovation Committee (Project JCYJ20150401145529049). We would like to thank Yin Zheng and Leo Yu Zhang for their valuable help.
\ifCLASSOPTIONcaptionsoff
  \newpage
\fi



%
\bibliographystyle{IEEEtran}
\bibliography{aaai18}

\begin{thebibliography}{10}
\providecommand{\url}[1]{#1}
\csname url@samestyle\endcsname
\providecommand{\newblock}{\relax}
\providecommand{\bibinfo}[2]{#2}
\providecommand{\BIBentrySTDinterwordspacing}{\spaceskip=0pt\relax}
\providecommand{\BIBentryALTinterwordstretchfactor}{4}
\providecommand{\BIBentryALTinterwordspacing}{\spaceskip=\fontdimen2\font plus
\BIBentryALTinterwordstretchfactor\fontdimen3\font minus
  \fontdimen4\font\relax}
\providecommand{\BIBforeignlanguage}[2]{{%
\expandafter\ifx\csname l@#1\endcsname\relax
\typeout{** WARNING: IEEEtran.bst: No hyphenation pattern has been}%
\typeout{** loaded for the language `#1'. Using the pattern for}%
\typeout{** the default language instead.}%
\else
\language=\csname l@#1\endcsname
\fi
#2}}
\providecommand{\BIBdecl}{\relax}
\BIBdecl

\bibitem{kulis2009fastCBIR}
B.~Kulis, P.~Jain, and K.~Grauman, ``Fast similarity search for learned
  metrics,'' \emph{IEEE Transactions on Pattern Analysis and Machine
  Intelligence}, vol.~31, no.~12, pp. 2143--2157, 2009.

\bibitem{gong2013iterativeCBIR}
Y.~Gong, S.~Lazebnik, A.~Gordo, and F.~Perronnin, ``Iterative quantization: A
  procrustean approach to learning binary codes for large-scale image
  retrieval,'' \emph{IEEE Transactions on Pattern Analysis and Machine
  Intelligence}, vol.~35, no.~12, pp. 2916--2929, 2013.

\bibitem{li2014subCBIR}
Y.~Li, C.~Chen, W.~Liu, and J.~Huang, ``Sub-selective quantization for
  large-scale image search.'' in \emph{AAAI}, 2014, pp. 2803--2809.

\bibitem{wang2017survey}
J.~Wang, T.~Zhang, N.~Sebe, H.~T. Shen \emph{et~al.}, ``A survey on learning to
  hash,'' \emph{IEEE Transactions on Pattern Analysis and Machine
  Intelligence}, 2017, in press.

\bibitem{HOG}
N.~Dalal and B.~Triggs, ``Histograms of oriented gradients for human
  detection,'' in \emph{Proceedings of the IEEE Computer Society Conference on
  Computer Vision and Pattern Recognition (CVPR)}, vol.~1, 2005, pp. 886--893.

\bibitem{gist}
A.~Oliva and A.~Torralba, ``Modeling the shape of the scene: A holistic
  representation of the spatial envelope,'' \emph{International Journal of
  Computer Vision}, vol.~42, no.~3, pp. 145--175, 2001.

\bibitem{SIFT}
D.~G. Lowe, ``Object recognition from local scale-invariant features,'' in
  \emph{Proceedings of the IEEE Conference on Computer Vision and Pattern
  Recognition (CVPR)}, vol.~2, 1999, pp. 1150--1157.

\bibitem{ke2004pcasift}
Y.~Ke and R.~Sukthankar, ``{PCA-SIFT}: A more distinctive representation for
  local image descriptors,'' in \emph{Proceedings of the 2004 IEEE Computer
  Society Conference on Computer Vision and Pattern Recognition (CVPR)},
  vol.~2, 2004, pp. 506--513.

\bibitem{GSIFT}
T.~Lindeberg, ``Image matching using generalized scale-space interest points,''
  \emph{Journal of Mathematical Imaging and Vision}, vol.~52, no.~1, pp. 3--36,
  2015.

\bibitem{shen2015supervisedhashing}
F.~Shen, C.~Shen, W.~Liu, and H.~Tao~Shen, ``Supervised discrete hashing,'' in
  \emph{Proceedings of the IEEE Conference on Computer Vision and Pattern
  Recognition (CVPR)}, 2015, pp. 37--45.

\bibitem{lin2014fastH}
G.~Lin, C.~Shen, Q.~Shi, A.~van~den Hengel, and D.~Suter, ``Fast supervised
  hashing with decision trees for high-dimensional data,'' in \emph{Proceedings
  of the IEEE Conference on Computer Vision and Pattern Recognition (CVPR)},
  2014, pp. 1963--1970.

\bibitem{kang2016column}
W.-C. Kang, W.-J. Li, and Z.-H. Zhou, ``Column sampling based discrete
  supervised hashing.'' in \emph{AAAI}, 2016, pp. 1230--1236.

\bibitem{CIFAR10}
A.~Krizhevsky and G.~Hinton, ``Learning multiple layers of features from tiny
  images,'' 2009, {U}niversity of Toronto Technical Report.

\bibitem{ILSVRC15}
O.~Russakovsky, J.~Deng, H.~Su, J.~Krause, S.~Satheesh, S.~Ma, Z.~Huang,
  A.~Karpathy, A.~Khosla, M.~Bernstein, A.~C. Berg, and L.~Fei-Fei, ``{ImageNet
  large scale visual recognition challenge},'' \emph{International Journal of
  Computer Vision}, vol. 115, no.~3, pp. 211--252, 2015.

\bibitem{chua2009nus}
T.-S. Chua, J.~Tang, R.~Hong, H.~Li, Z.~Luo, and Y.~Zheng, ``{NUS-WIDE}: A
  real-world web image database from {N}ational {U}niversity of {S}ingapore,''
  in \emph{Proceedings of the ACM International Conference on Image and Video
  Retrieval}, 2009, article no. 48.

\bibitem{coco}
T.-Y. Lin, M.~Maire, S.~Belongie, J.~Hays, P.~Perona, D.~Ramanan,
  P.~Doll{\'a}r, and C.~L. Zitnick, ``Microsoft {COCO}: {C}ommon objects in
  context,'' in \emph{European Conference on Computer Vision}, 2014, pp.
  740--755.

\bibitem{DPSH}
W.-J. Li, S.~Wang, and W.-C. Kang, ``Feature learning based deep supervised
  hashing with pairwise labels,'' in \emph{Proceedings of the International
  Joint Conference on Artificial Intelligence (IJCAI)}, 2016, pp. 1711--1717.

\bibitem{2017hashNet}
Z.~{Cao}, M.~{Long}, J.~{Wang}, and P.~S. {Yu}, ``{HashNet: Deep learning to
  hash by continuation},'' \emph{ArXiv preprint arXiv:1702.00758}, 2017.

\bibitem{xia2014supervised}
R.~Xia, Y.~Pan, H.~Lai, C.~Liu, and S.~Yan, ``Supervised hashing for image
  retrieval via image representation learning.'' in \emph{AAAI}, vol.~1, 2014,
  pp. 2156--2162.

\bibitem{DTSH}
X.~Wang, Y.~Shi, and K.~M. Kitani, ``Deep supervised hashing with triplet
  labels,'' in \emph{Asian Conference on Computer Vision}, 2016, pp. 70--84.

\bibitem{facenet}
F.~Schroff, D.~Kalenichenko, and J.~Philbin, ``Facenet: {A} unified embedding
  for face recognition and clustering,'' in \emph{the IEEE Conference on
  Computer Vision and Pattern Recognition (CVPR)}, 2015, pp. 815--823.

\bibitem{yang2017superviseddeepbinary}
H.-F. Yang, K.~Lin, and C.-S. Chen, ``Supervised learning of
  semantics-preserving hash via deep convolutional neural networks,''
  \emph{IEEE Transactions on Pattern Analysis and Machine Intelligence}, 2017,
  in press.

\bibitem{yao2016}
T.~Yao, F.~Long, T.~Mei, and Y.~Rui, ``Deep semantic-preserving and
  ranking-based hashing for image retrieval.'' in \emph{Proceedings of the
  International Joint Conference on Artificial Intelligence (IJCAI)}, 2016, pp.
  3931--3937.

\bibitem{yansemi}
X.~Yan, L.~Zhang, and W.-J. Li, ``Semi-supervised deep hashing with a bipartite
  graph,'' \emph{Proceedings of the Twenty-Sixth International Joint Conference
  on Artificial Intelligence (IJCAI)}, pp. 3238--3244, 2017.

\bibitem{MagnetRippel2016}
O.~Rippel, M.~Paluri, P.~Dollar, and L.~Bourdev, ``Metric learning with
  adaptive density discrimination,'' \emph{arXiv preprint arXiv:1511.05939v1},
  2015.

\bibitem{zhuang2016fasttriplet}
B.~Zhuang, G.~Lin, C.~Shen, and I.~Reid, ``Fast training of triplet-based deep
  binary embedding networks,'' in \emph{Proceedings of the IEEE Conference on
  Computer Vision and Pattern Recognition (CVPR)}, 2016, pp. 5955--5964.

\bibitem{lai2015DNNH}
H.~Lai, Y.~Pan, Y.~Liu, and S.~Yan, ``Simultaneous feature learning and hash
  coding with deep neural networks,'' in \emph{Proceedings of the IEEE
  Conference on Computer Vision and Pattern Recognition (CVPR)}, 2015, pp.
  3270--3278.

\bibitem{DSDH}
Q.~Li, Z.~Sun, R.~He, and T.~Tan, ``Deep supervised discrete hashing,''
  \emph{arXiv preprint arXiv:1705.10999}, 2017.

\bibitem{SSDH}
H.-F. Yang, K.~Lin, and C.-S. Chen, ``Supervised learning of
  semantics-preserving hash via deep convolutional neural networks,''
  \emph{IEEE Transactions on Pattern Analysis and Machine Intelligence}, 2017,
  in press.

\bibitem{yao2016DSRH}
T.~Yao, F.~Long, T.~Mei, and Y.~Rui, ``Deep semantic-preserving and
  ranking-based hashing for image retrieval.'' in \emph{Proceedings of the
  International Joint Conference on Artificial Intelligence (IJCAI)}, 2016, pp.
  3931--3937.

\bibitem{cui2016fine}
Y.~Cui, F.~Zhou, Y.~Lin, and S.~Belongie, ``Fine-grained categorization and
  dataset bootstrapping using deep metric learning with humans in the loop,''
  in \emph{Proceedings of the IEEE Computer Society Conference on Computer
  Vision and Pattern Recognition (CVPR)}, 2016, pp. 1153--1162.

\bibitem{personReID}
D.~Cheng, Y.~Gong, S.~Zhou, J.~Wang, and N.~Zheng, ``Person re-identification
  by multi-channel parts-based cnn with improved triplet loss function,'' in
  \emph{Proceedings of the IEEE Conference on Computer Vision and Pattern
  Recognition (CVPR)}, 2016, pp. 1335--1344.

\bibitem{QuintupletHuang2016}
C.~Huang, Y.~Li, C.~Change~Loy, and X.~Tang, ``Learning deep representation for
  imbalanced classification,'' in \emph{Proceedings of the IEEE Conference on
  Computer Vision and Pattern Recognition (CVPR)}, 2016, pp. 5375--5384.

\bibitem{liftedStructure}
H.~Oh~Song, Y.~Xiang, S.~Jegelka, and S.~Savarese, ``Deep metric learning via
  lifted structured feature embedding,'' in \emph{Proceedings of the IEEE
  Computer Society Conference on Computer Vision and Pattern Recognition
  (CVPR)}, 2016, pp. 4004--4012.

\bibitem{yang2017pairwise}
E.~Yang, C.~Deng, W.~Liu, X.~Liu, D.~Tao, and X.~Gao, ``Pairwise relationship
  guided deep hashing for cross-modal retrieval.'' in \emph{AAAI}, 2017, pp.
  1618--1625.

\bibitem{zhu2016DHN}
H.~Zhu, M.~Long, J.~Wang, and Y.~Cao, ``Deep hashing network for efficient
  similarity retrieval.'' in \emph{AAAI}, 2016, pp. 2415--2421.

\bibitem{ioffe2015batch}
S.~Ioffe and C.~Szegedy, ``Batch normalization: Accelerating deep network
  training by reducing internal covariate shift,'' in \emph{International
  Conference on Machine Learning}, 2015, pp. 448--456.

\bibitem{HashClusteringong2015}
Y.~Gong, M.~Pawlowski, F.~Yang, L.~Brandy, L.~Bourdev, and R.~Fergus, ``Web
  scale photo hash clustering on a single machine,'' in \emph{Proceedings of
  the IEEE Conference on Computer Vision and Pattern Recognition (CVPR)}, 2015,
  pp. 19--27.

\bibitem{chen2015mxnet}
T.~Chen, M.~Li, Y.~Li, M.~Lin, N.~Wang, M.~Wang, T.~Xiao, B.~Xu, C.~Zhang, and
  Z.~Zhang, ``Mxnet: {A} flexible and efficient machine learning library for
  heterogeneous distributed systems,'' \emph{arXiv preprint arXiv:1512.01274},
  2015.

\bibitem{DSHNP}
X.~C. R.~B. Sen~Su, Gang~Chen, ``Deep supervised hashing with nonlinear
  projections,'' in \emph{Proceedings of the International Joint Conference on
  Artificial Intelligence (IJCAI)}, 2017, pp. 2786--2792.

\bibitem{DQN}
Y.~Cao, M.~Long, J.~Wang, H.~Zhu, and Q.~Wen, ``Deep quantization network for
  efficient image retrieval.'' in \emph{AAAI}, 2016, pp. 3457--3463.

\bibitem{jarvelin2002cumulated}
K.~J{\"a}rvelin and J.~Kek{\"a}l{\"a}inen, ``Cumulated gain-based evaluation of
  {IR} techniques,'' \emph{ACM Transactions on Information Systems}, vol.~20,
  no.~4, pp. 422--446, 2002.

\bibitem{wang2017Hierarchical}
D.~Wang, H.~Huang, C.~Lu, B.-S. Feng, L.~Nie, G.~Wen, and X.-L. Mao,
  ``Supervised deep hashing for hierarchical labeled data,'' \emph{arXiv
  preprint arXiv:1704.02088}, 2017.

\bibitem{Center_lossWen2016}
Y.~Wen, K.~Zhang, Z.~Li, and Y.~Qiao, ``A discriminative feature learning
  approach for deep face recognition,'' in \emph{European Conference on
  Computer Vision}, 2016, pp. 499--515.

\bibitem{liu2017coco_loss}
Y.~Liu, H.~Li, and X.~Wang, ``Learning deep features via congenerous cosine
  loss for person recognition,'' in \emph{Advances in Neural Information
  Processing Systems (NIPS)}, 2017.

\bibitem{maaten2008t-SNE}
L.~Maaten and G.~Hinton, ``Visualizing data using t-{SNE},'' \emph{Journal of
  Machine Learning Research}, vol.~9, no. Nov., pp. 2579--2605, 2008.

\bibitem{n-pairs}
K.~Sohn, ``Improved deep metric learning with multi-class n-pair loss
  objective,'' in \emph{Advances in Neural Information Processing Systems
  (NIPS)}, 2016, pp. 1857--1865.

\end{thebibliography}

%

\begin{IEEEbiography}[{\includegraphics[width=1in,height=1.25in,clip,keepaspectratio]{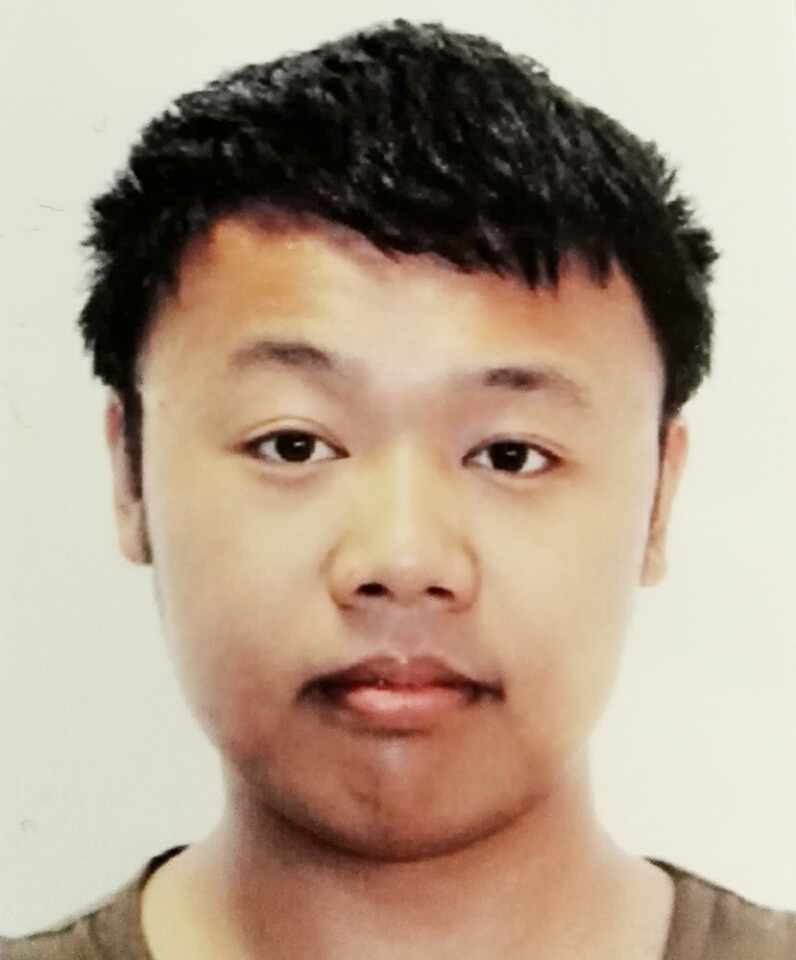}}]{\textbf{Xuefei Zhe} recevied the BSc degree in Information Engineering from Nanjing University, China in 2014. He is currently working toward the PhD degree in the Department of Electronic Engineering, City University of Hong Kong. His research interests are in computer vision and deep learning.}
\end{IEEEbiography}


\begin{IEEEbiography}[{\includegraphics[width=1in,height=1.25in,clip,keepaspectratio]{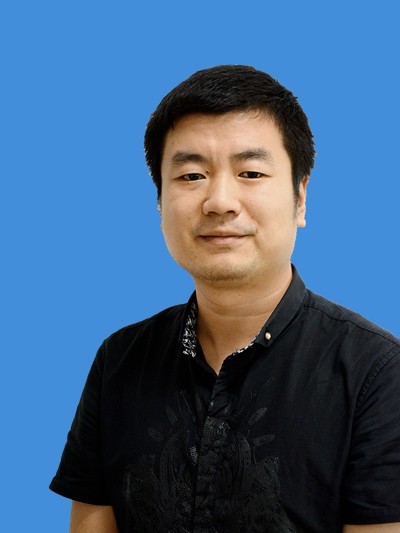}}]{\textbf{Shifeng Chen} received the B.E. degree from the University of Science and Technology of China, Hefei, in 2002, the M.Phil. degree from City University of Hong Kong, Hong Kong, in 2005, and the Ph.D. Degree from the Chinese University of Hong Kong, Hong Kong, in 2008. He is now an Associate Professor in the Shenzhen Institutes of Advanced Technology, Chinese Academy of Sciences, China. His research interests include computer vision and machine learning.}
\end{IEEEbiography}

\begin{IEEEbiography}[{\includegraphics[width=1in,height=1.25in,clip,keepaspectratio]{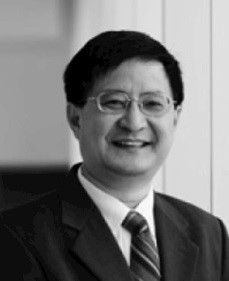}}]{\textbf{Hong Yan} received his Ph.D. degree from Yale University. He was professor of imaging science at the University of Sydney and currently is professor of computer engineering at City University of Hong Kong. His research interests include image processing, pattern recognition and bioinformatics. He has authored or co-authored over 300 journal and conference papers in these areas. He was elected an IAPR fellow for contributions to document image analysis and an IEEE fellow for contributions to image recognition techniques and applications.}
\end{IEEEbiography}



\newpage
\section{Supplementary Material}

\subsection{Top 21 tags of NUS-WIDE}
The top 21 tags of NUS-WIDE are: mountain beach tree snow vehicle rocks reflection sunset flowers road ocean lake plants window buildings grass animal water person clouds sky.
\subsection{Retrieval Results}
\begin{figure*}[h]
    \centering
    \includegraphics[width=0.7\textwidth]{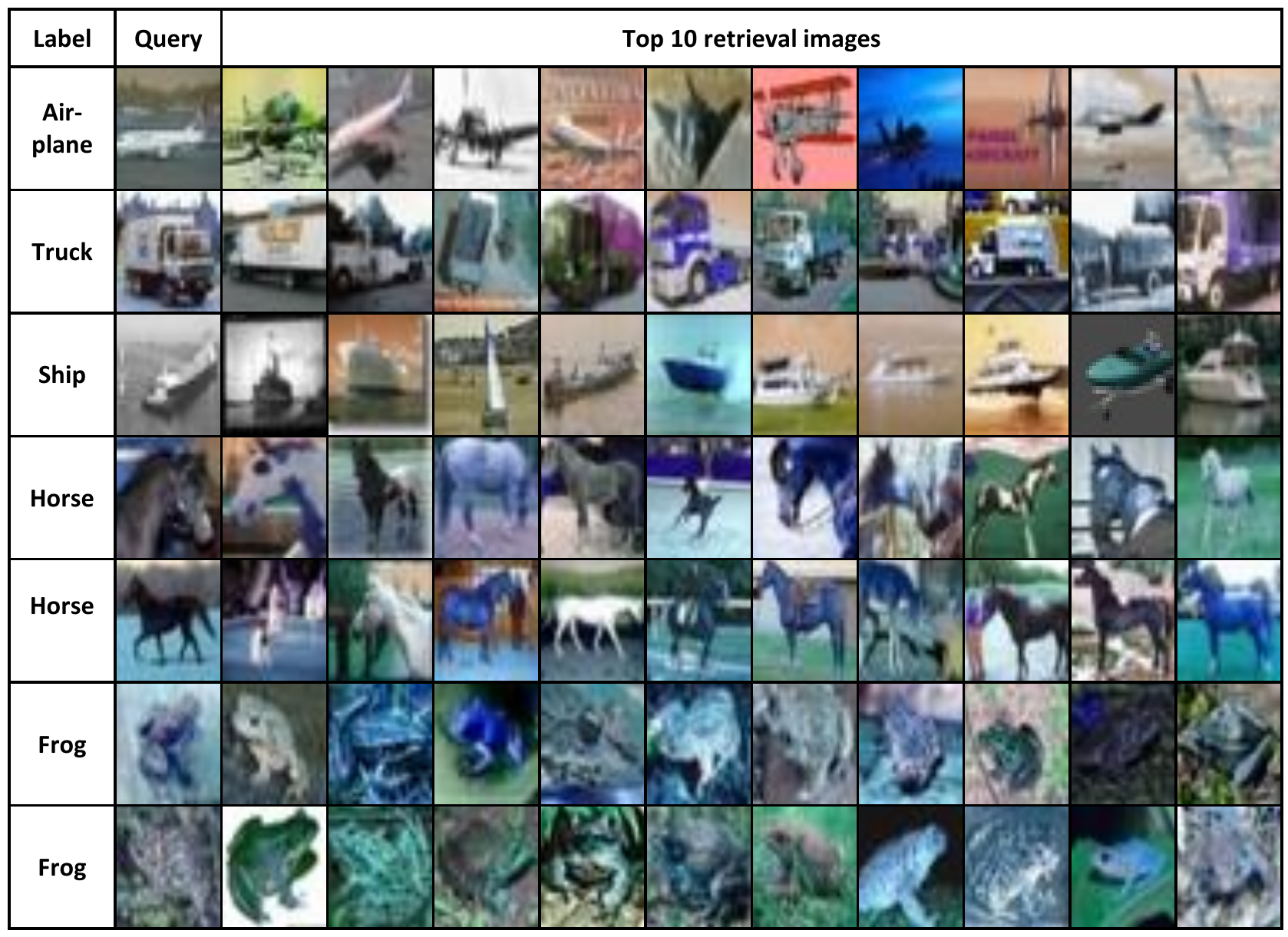}
    \caption{Retrieval results of CIFAR-10}
    \label{fig:cifar_r}
\end{figure*}

\begin{figure*}[b!]
    \centering
    \includegraphics[width=0.8\textwidth]{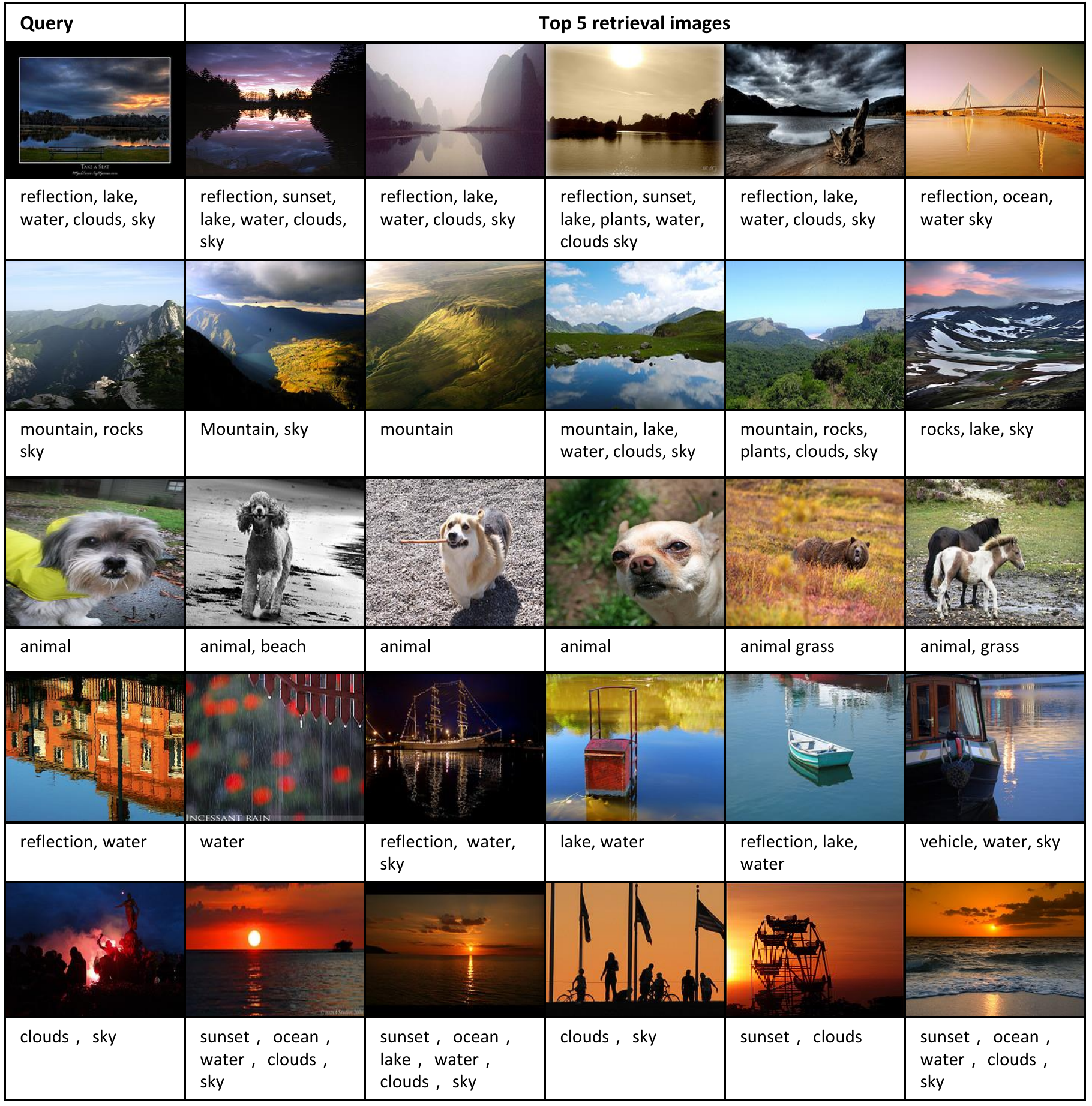}
    \caption{Retrieval results of NUS-WIDE}
    \label{fig:nus_res}
\end{figure*}

\begin{figure*}[b!]
    \centering
    \includegraphics[width=0.9\textwidth]{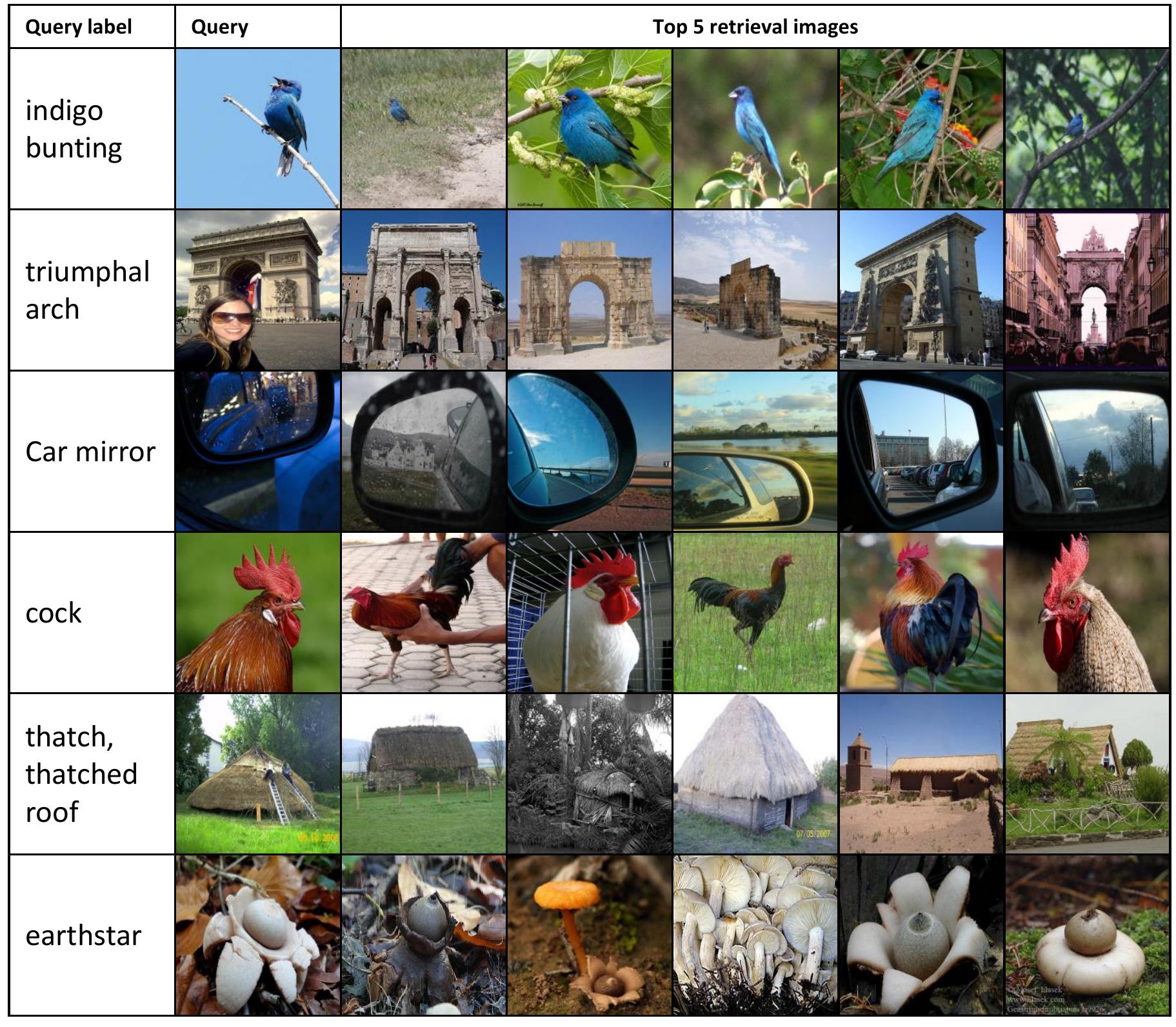}
    \caption{Retrieval results of ImageNet-100}
    \label{fig:imgNet-100_r}
\end{figure*}

We presents more retrieval results of CIFAR-10, NUS-WIDE and ImageNet-100 in Figure~\ref{fig:cifar_r}, Figure~\ref{fig:nus_res} and Figure~\ref{fig:imgNet-100_r}.

\end{document}